\documentclass[10pt,twocolumn,letterpaper]{article}
\usepackage{eso-pic}
\usepackage{cvpr}
\usepackage{times}
\usepackage{epsfig}
\usepackage{graphicx}
\usepackage{amsmath}
\usepackage{amssymb}
\usepackage{booktabs}
\usepackage{comment}
\usepackage{capt-of,etoolbox}
\usepackage{tabularx}

\newcommand{\lmss}[1]{\fontfamily{lmss}{\selectfont #1}}
\usepackage{multirow, makecell}
\usepackage{colortbl}
\definecolor{Gray}{gray}{0.9}

\usepackage[pagebackref=true,breaklinks=true,letterpaper=true,colorlinks,bookmarks=false]{hyperref}

\cvprfinalcopy %

\def\cvprPaperID{1798} %

\ifcvprfinal\fi

\begin{document}

\makeatletter
\apptocmd\@maketitle{{\myfigure{}\par}}{}{}
\makeatother

\newcommand\myfigure{
\vspace{-2em}
\centering
    \includegraphics[width=0.98\linewidth]{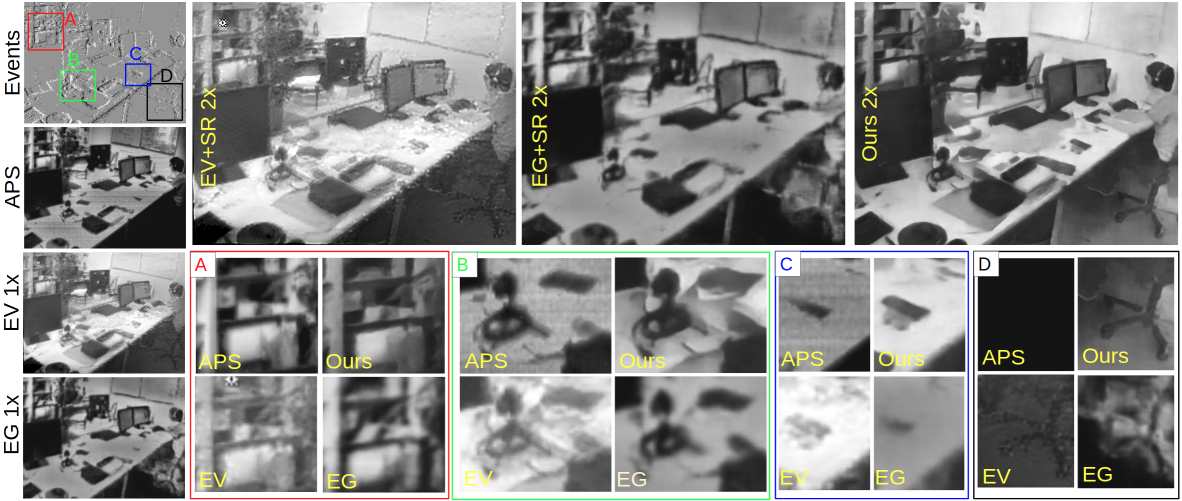}
\captionof{figure}{Reconstructing high-definition photo-realistic intensity images from pure events in end-to-end learning. Our events to super-resolved intensity image reconstruction recovers more details with less artifacts in comparison to recent methods of EG~\cite{mostafavi2019event} and EV~\cite{rebecq2019high}.
}
\label{teaser}
\vspace{15 pt}
}

\title{Learning to Super Resolve Intensity Images from Events\\
{\small \textbf{\url{https://github.com/gistvision/e2sri}}}
\vspace{-0.5em}
}

\author{
S. Mohammad Mostafavi I.\\
GIST, South Korea\\
{\tt\small mostafavi@gist.ac.kr}
\and
Jonghyun Choi\\
GIST, South Korea\\
{\tt\small jhc@gist.ac.kr}
\and
Kuk-Jin Yoon\\
KAIST, South Korea\\
{\tt\small kjyoon@kaist.ac.kr}\\
}

\maketitle

\begin{abstract}
An event camera detects per-pixel intensity difference and produces asynchronous event stream with low latency, high dynamic range, and low power consumption.  As a trade-off, the event camera has low spatial resolution. We propose an end-to-end network to reconstruct high resolution, high dynamic range (HDR) images directly from the event stream.  We evaluate our algorithm on both simulated and real-world sequences and verify that it captures fine details of a scene and outperforms the combination of the state-of-the-art event to image algorithms with the state-of-the-art super resolution schemes in many quantitative measures by large margins. We further extend our method by using the active sensor pixel (APS) frames or reconstructing images iteratively.
\end{abstract}

\vspace{-1em}
\section{Introduction}

Event cameras, also known as neuromorphic cameras, have successfully opened their path to the computer vision and robotics society for its low cost and high dynamic sensing range with low latency and low power consumption. 
It represents the changes of intensity for a pixel location $(x,y)$ as a plus or minus sign $(\sigma)$ asynchronously by checking the amount of intensity changes with a predefined threshold. 
This stream-like representation, depending on the scene and camera movement, can achieve $\mu$s order of latency through accurate timestamps $(t)$ and is expressed per fired event in the form of $(x,y,t,\sigma)$. 
This device has garnered a lot of attention due to the high applicability in systems requiring high dynamic range outputs with low latency, and low power and low memory consumption constraints \cite{mueggler2017event,mostafavi2019event,rebecq2019high,zhu2018multivehicle,Gehrig_2019_ICCV}.  
New applications for the event cameras have emerged such as intensity image reconstruction or recovering geometric features such as optical flow or depth from the event stream \cite{bardow2016simultaneous,reinbacher2016real,kim2008simultaneous,cook2011interacting,scheerlinck2018continuous,tulyakov2019learning}. 

Unfortunately, most commercially available event cameras produce relatively low resolution event streams for their efficiency.
While there are number of proposals on many applications%
estimating super-resolved intensity images from the events has been barely explored in the literature.
To generate the high resolution images from the event, one can combine a method to transfer events to intensity images with a super resolution algorithm for intensity images \cite{dai2019second,sajjadi2018frame,haris2019recurrent, lim2017enhanced}.
But these pipelined approaches are sub-optimal in generating the high resolution images from the events and may fail to reconstruct details of scenes.
For producing high fidelity high resolution images, we aim to directly learn to estimate pixel-wise super-resolved intensity from events in an end-to-end manner and demonstrate that our method is able to super resolve images with rich details and less artifacts, better than pipelined state of the arts in both qualitative and quantitative analyses.

To the best of our knowledge, we are the first to model super-resolving event data to higher-resolution intensity images in an end-to-end learning framework. 
We further extend our method to reconstruct more details by considering APS frames as inputs or learning the network iteratively to add details to an initial image.

\section{Related Work}

\paragraph{Event to intensity images.}
Early attempts in the applications of event cameras, consider relatively short periods of the event stream data and direct accumulation of the plus or minus events in two colors as a gradient interpreted output \cite{brandli2014240}.
Synthesising intensity images instead of the gradient representation is originated from the task of simultaneously estimating the camera movement and mosaicing them as a panoramic gradient image \cite{kim2008simultaneous}. 
In their approach the scene is static and the camera only has rotational movements. By the Poisson integration they transfer a gradient image to an intensity image. 
In \cite{cook2011interacting}, a bio-inspired network structure of recurrently interconnected maps is proposed to predict different visual aspects of a scene such as intensity image, optical flow, and angular velocity from small rotation movements. 
In \cite{bardow2016simultaneous}, a joint estimation of optical flow and intensity simultaneously in a variational energy minimization scheme in a challenging dynamic movement setting is proposed. However, their method propagates errors as shadow-like artifacts in the generated intensity images. 

A variational framework based on a denoising scheme that filters incoming events iteratively is introduced in \cite{reinbacher2016real}. They utilized manifold regularization on the relative timestamp of events to reconstruct the image with more grayscale variations in untextured areas.
In \cite{scheerlinck2018continuous}, an asynchronous high-pass filter is proposed to reconstruct videos in a computationally efficient manner. 
This framework is originally designed for complementing intensity frames with the event information but is also capable of reconstructing images from events without the help of APS frames.

Recent approaches use deep convolutional networks to create photo-realistic images directly from the event stream \cite{mostafavi2019event,rebecq2019high}. Both approaches employ a $U{\text -}net$ \cite{ronneberger2015u} as their base architecture with modifications such as using conditional generative adversarial neural networks \cite{mostafavi2019event} or using a deep recurrent structure (up to 40 steps) together with stacked ConvLSTM gates \cite{rebecq2019high}. They further investigated the possibility of reaching very high frame rates and using the output intensity images for downstream applications.

\vspace{-1em}\paragraph{Image super resolution (SR).}
Intensity image SR algorithms can be largely categorized into single image SR (SISR) \cite{dai2019second, lim2017enhanced} or multiple image SR (MISR) also known as video SR \cite{sajjadi2018frame,haris2019recurrent}.
SISR methods add details inferred from the context of the given single low resolution (LR) image while MISR further uses a sequence of images over time. 
Since MISR uses more LR images to reconstruct the high resolution image, it is generally more successful in recovering missing details and higher frequency information. 
Since we have a sequence of stacks, MISR is more similar to our approach, although we aim to reconstruct one single image each time. 
The learning based SR methods outperform previous methods by using deeper and wider networks while utilizing the power of residual connections to prevent vanishing gradients \cite{lim2017enhanced,haris2019recurrent}. 
Many MISR methods use optical flow representations among the input images as a supplementary source of input to reach higher quality SR outputs \cite{sajjadi2018frame,haris2019recurrent}. 
Inspired by these methods, we design our SR sub-network as described in Sec. \ref{SR_network}.

\section{Approach}

\begin{figure}[t]
\centering
\includegraphics[width=1\linewidth]{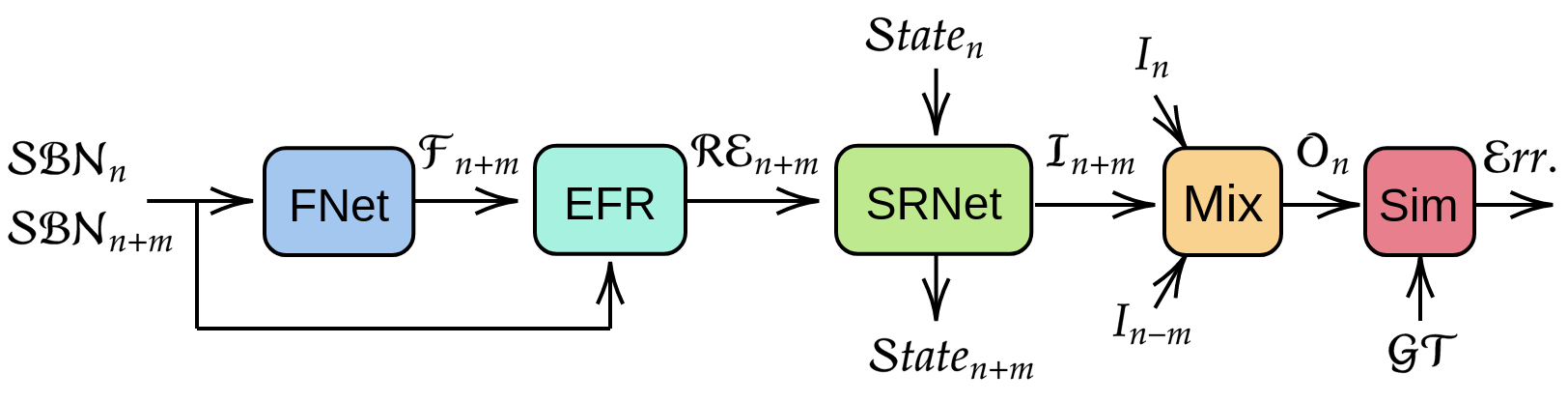}
\caption{Overview of our end-to-end event to super-resolved intensity image framework. The input stacks $SBN_{n+m}$ and the central stack $SBN_n$ are given to the FNet to create the optical flow ($F_{n+m}$). The flow and stacks are concatenated and given to the EFR to rectify the event features. Its output $RE_{n+m}$ is given to SRNet together with the previous state ($State_n$) to create intermediate intensity outputs $I_{n+m}$ and the next state ($State_{n+m}$). All intermediate intensity outputs are concatenated and given to the mixer ($Mix$) network which creates the final output ($O_n$). Finally, the output is compared to the training groundtruth (GT) using the similarity loss (Sim) including Learned Perceptual Image Patch Similarity (LPIPS) term and $\ell_1$ term to compute error ($Err$).}
\label{fig:flow}
\vspace{-1em}
\end{figure}

\begin{figure*}[t]
\centering
\vspace{5 pt}
 \includegraphics[width=1\linewidth]{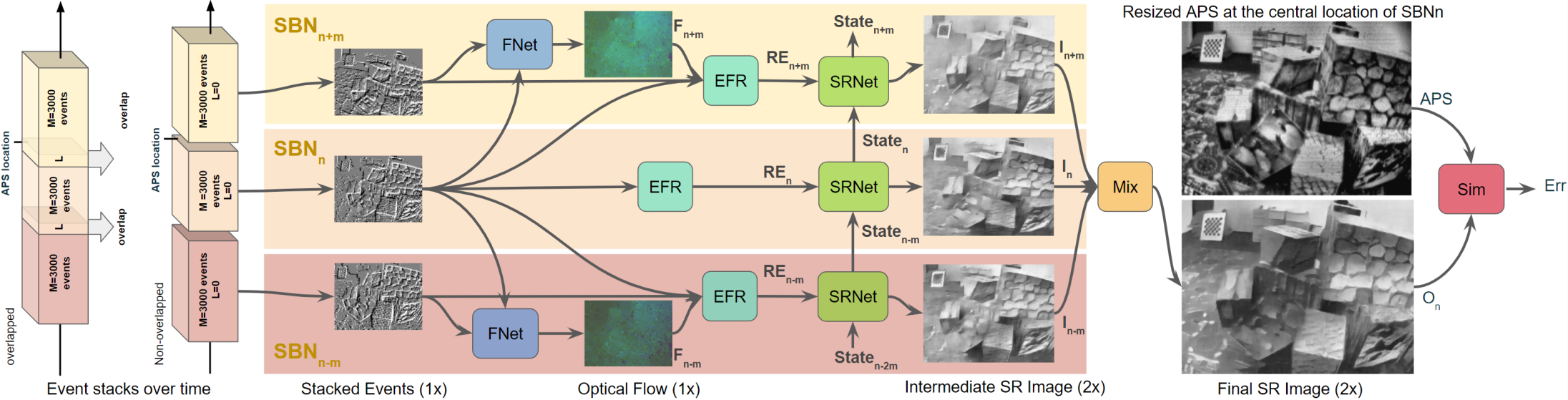}
\caption{Detailed data flow in the proposed method. This example is based on third stack ($SBN_{n+m}$), therefore the previous inputs, optical flow, and intermediate intensity outputs are faded. The APS frame is resized to the size of output ($O_n$) for comparison}
\vspace{-1em}
\label{interm}
\end{figure*}

We propose a fully convolutional network that takes a sequence of events stacks near the timestamp of interest as input, relates them in pairs with their optical flow obtained by $FNet$ and rectify the combination of the paired stacks and the flow by $EFR$, then feeds them to the recurrent neural network based super-resolution network ($SRNet$) that outputs hidden states and intermediate intensity outputs per each stack. Finally, we mix the intermediate outputs from multiple time stamps by $Mix$ to construct a super resolved intensity image.
We briefly illustrate the structure in Fig. \ref{fig:flow} and with the detailed data flow in  Fig. \ref{interm} in Sec. \ref{sec:overall_structure}.
Beginning with event stacking strategy, we describe the details of our network architecture. %

\subsection{Event Stacking Method}
\label{event_prepare}
The stream-like representation of events is sparse in spatial domain and needs preparation to capture scene details to be reconstructed by a convolutional neural network.
Despite recent advances of the stacking methods ~\cite{Gehrig_2019_ICCV,tulyakov2019learning}, our network performs well with a simple stacking method such as \emph{stacking based on the number of events} (SBN) \cite{mostafavi2019event}. 
Employing the advanced stacking methods is straightforward by minor modifications to %
the input blocks of our network.

With the SBN, starting from any timestamp in the event stream, we count the number of events until we reach a predefined number ($N_{e}$) and accumulate the events to form one \emph{channel} in the stack. 
We repeat this process $c$ times for one \emph{stack}. 
Thus, each stack contains $M = c \times N_e$ events in total and has the dimension of $h{\times}w{\times}c$, where $h$ and $w$ are the width and height of the APS images, respectively.
This $c$-channel stack is fed into the network as an input. 
The corresponding APS frame is sampled at the timestamp of the last event in the stack for the ground truth (GT).
At each channel, all pixel values are initially set to $128$. If an event is triggered at location $(x,y)$, we replace the pixel value at $(x,y)$ in the same channel with $256$ (positive event) or $0$ (positive event). 
Since newly coming events can override older events, the $M$ needs to be carefully chosen to better preserve spatio-temporal visual information.
The frame rate can be determined by both the $N_e$ and the number of overlapping events between each stack over time.

We empirically choose to use $3,000$ events per stack in which each stack has %
3 channels.
This number can be modified for the experiments with larger resolution event inputs to ensure that the average number of events in stacks %
show visually plausible outputs with fine details.
However, since the network is trained on diverse scenes which contain different numbers of local events, the network is not very sensitive to the chosen number of events per stack at inference.

\vspace{-0.2em}
\subsection{Network Architecture}
\label{net_arch}
We design the network architecture by three principles. 
First, we take into account the characteristics of the input and target (Sec.~\ref{sec:overall_structure}). 
Second, we have a sufficiently large hypothesis space for the super-resolution network ($SRNet$) to address various level of complexity of movements in a scene (Sec.~\ref{SR_network}).
Finally, we propose a novel objective function that can add structural details while being away from blur and artifacts (Sec.~\ref{criterion}). 
We %
describe the details of each component of our proposed network. %

\vspace{-6 pt}
\subsubsection{Overview}
\label{sec:overall_structure}

We consider a stream of events stacked for the input to our network. 
In particular, for the input sequence of three stacks ($3S$), the stacks are the one containing the $n^\text{th}$ APS timestamp ($SBN_n$), the stack before it $SBN_{n-m}$ and the stack after it ($SBN_{n+m}$).
We illustrate the network with these inputs in Fig. \ref{interm} with detailed data flow through the sub-networks.
Note that the network can be used with the input of any number of stacks in a sequence (\eg, 3 or 7).

Each stack has $M$ (\eg, $3,000$) events and its end location $m$ will vary on the timeline of events based on the amount of time it is required to fire $M$ events. $SBN_n$ is the \emph{central stack} among the three sequences. 
It is fed to the network after $SBN_{n-m}$ and the predicted intensity output is corresponding to this stack. 
The $SBN_{n+m}$ and $SBN_{n-m}$ stacks are $M$ events away from the beginning or end of the central stack respectively if there is no overlap ($L=0$) among the stacks (`Non-overlapped' input in Fig. \ref{interm}). %
We can also have overlapping stacks for creating higher frame-rates; the end of the next stack will be $M$ events after the center minus the amount of overlap ($M {\scriptstyle-} L$) (`overlapped' input in Fig. \ref{interm}). More details 
on the overlapped stacking is provided in the supplement.

$SBN_{n+m}$ and $SBN_{n-m}$ are fed separately with the central stack to the optical flow estimation network ($FNet$) to predict the optical flow ($F_{n+m}$ or $F_{n-m}$) between the stacks. 
These stacks of events are concatenated with the optical flow obtained by the $FNet$ and then rectified by an event feature rectification network ($EFR$).
The rectified event stack ($RE_{n+m}$) is then given to the super-resolution network ($SRNet$). 
The $SRNet$ takes the previous state ($State_n$) with the rectified events stack ($RE_{n+m}$) and creates the next state ($State_{n+m}$) of the sequential model and a super-resolved intensity like output ($I_{n+m}$). 

Since the stacks quantize continuous event stream into separate inputs, each stack may not contain all necessary details for reconstructing images. 
Thus, the intermediate intensity outputs from all the stacks are then mixed by a Mixer network ($Mix$) to reconstruct intensity image $O_n$ with rich details. %
For the initial stack, only the first stack is fed to the $EFR$ sub-network to create an initial $State_n$.
The output of $Mix$ is given to the similarity network ($Sim$) to optimize the parameters based on the error ($Err$).

\subsubsection{Flow Network (FNet)}

An unwanted downside of stacking the event stream is losing temporal relation between the stacks. 
The lost temporal relation between stacks can be partially recovered by using a sequence of the stacks and the optical flow between each pair of stacks as the optical flow reports how the triggered events in the scene have moved and in which location the changes have happened. 
The SBN stacking includes sufficient edge information and can be used as an image-like input to well-known learning-based optical flow estimation algorithms.
Thus, we do not finetune it but use a pretrained $FNet$ for computational efficiency\footnote{Finetuning $FNet$ may further improve the output quality as the stacked image has different visual signature from natural images.}.
We use \cite{ilg2017flownet} as our flow estimation network and call it as $FNet$. %

\vspace{-6 pt} %
\subsubsection{Event Feature Rectification Network (EFR)}
Another downside of stacking events is overwriting previous event information in fast triggering locations. 
The overwritten events result in a blurry stack of events and eventually lower quality reconstructions. 
To prevent overwriting events, we concatenate two stack of events with the optical flow and provide it to two convolutional layers called the event feature rectification ($EFR$) network. 
By the $EFR$, we progressively fuse the stacks over the event stream to preserve details from each event.

The $EFR$ helps to reconstruction images when two stacks have events in a location visible to only one stack which the optical flow cannot relate, the events will more likely be maintained for the intensity reconstruction since we use all three inputs by the $EFR$.
Note that the central stack is provided to this network without estimated flow since there is no flow for it.

\vspace{-6 pt} %
\subsubsection{Super Resolution Network (SRNet)}
\label{SR_network}

The rectified events are now super resolved by our main network called $SRNet$.
We use a recurrent neural network for the $SRNet$ because each part of the event stream which we stack captures details of the output image and they are originally continuous but quantized by the stacking method.
To alleviate the discontinuity, we utilize the internal memory state of recurrent neural network to reconstruct different regions with rich details in a continuous manner as the state is updated internally by each incoming stack.
Specifically, a single event stack might partially miss important details from previously fired events which are not in its stacking range but have been captured by the previous stacks. 

It has been shown that stacked events are capable of synthesizing intensity images by deep neural networks \cite{mostafavi2019event,rebecq2019high} such as $U{\text -}net$ \cite{ronneberger2015u}. 
Architecturally, we further extend the idea by using $ResNet$ \cite{he2016deep} with 15 blocks in depth with more filters and larger kernel size.
In particular, following the well-designed networks in MISR \cite{lim2017enhanced,sajjadi2018frame,haris2018deep,dai2019second}, we utilize the power of residual learning for super-resolving intensity. 
We use large field of views inspired from the SISR network \cite{haris2018deep} to transfer the rectified event features to SR intensity generator ($RNet{\text -}C$). Its main task is to create an initial SR intensity image state by the combination of transposed convolutional operations. 

\begin{figure}[t!]
\centering
        \includegraphics[width=1\linewidth]{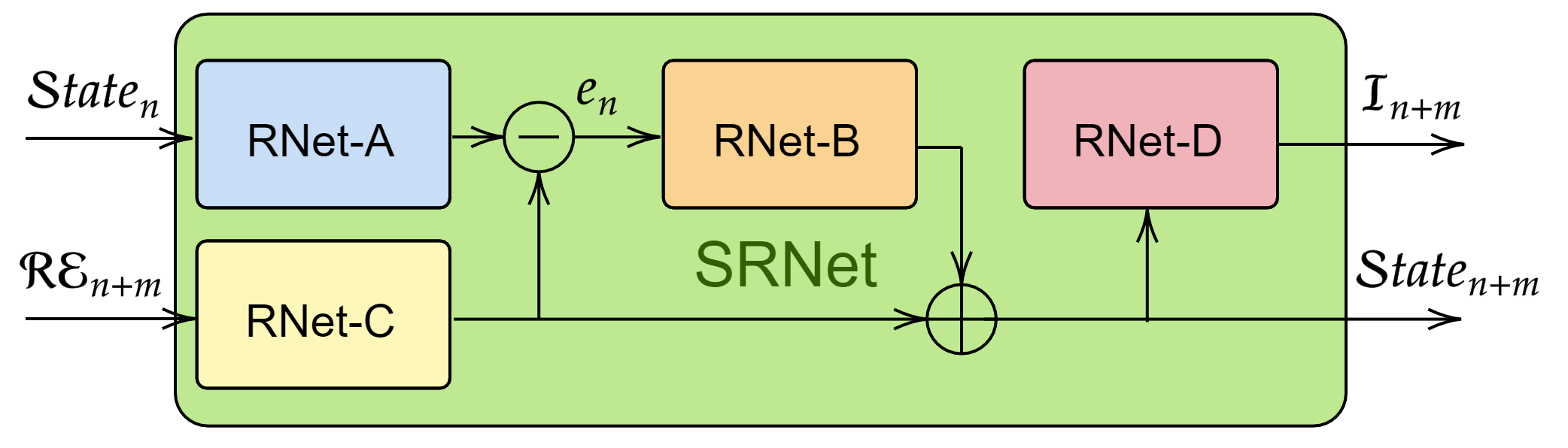}
\caption{Detailed architecture of the proposed super resolving network ($SRNet$) (Green-block in Fig.~\ref{fig:flow}). Four main residual networks are designed to perform as a large encoder-decoder scheme. $RNet{\text -}A$ is used to update the hidden state while $RNet{\text -}B$ and $RNet{\text -}D$ act as an encoder and decoder respectively to map the hidden state as a super resolved intensity output ($I_{n+m}$). 
}
\label{fig:SRNet}
\vspace{-10pt}
\end{figure}

The $SRNet$ is designed to upscale the input $RE$ while adding intensity information. 
The overall structure of the $SRNet$ is illustrated in Fig.~\ref{fig:SRNet}.
We use the combination of three residual networks ($RNet{-}\{A,B,D\}$) that are composed of five ResNet blocks containing two convolutional layers.
These networks are shallower than $RNet{\text -}C$ because they encode feature-like representations from previous states and not directly from the rectified events. %
The output of $RNet{\text -}A$ which performs as an upsampling encoder is subtracted from the output of $RNet{\text -}C$ to create an internal error ($e_n$), which measures how much the current rectified event stack $RE_{n+m}$ contributes in comparison to the previous state $State_n$ as
\begin{equation}
e_n = RNet{\text -}C(RE_{n+m}) - RNet{\text -}A(State_n).
\label{eq1}
\end{equation}
This error is given as an input to $RNet{\text -}B$ which performs as a general encoder. 
We define the the next state ($State_{n+m}$) by the output of $RNet{\text -}B$ summed with $RNet{\text -}C$ thus the current input ($RE_{n+m}$) is emphasized as
\begin{equation}
State_{n+m} = RNet{\text -}B(e_n) + RNet{\text -}C(RE_{n+m}).
\end{equation}
The $State_{n+m}$ is given to a final decoder ($RNet{\text -}D$) to make the intermediate intensity output ($I_{n+m}$) as
\begin{equation}
I_{n+m} = RNet{\text -}D(State_{n+m}).
\end{equation}
In general, the $RNet{\text -}C$ adds new information from the current stack to the previous state by adding details of the scene missed by the previous stack.
Even when there is no events in some regions captured by the current stack but there are scene details in the regions captured by the previous stack, the previous state ($State_n$) holds that information through $RNet{\text -}A$ as its hidden state to reconstruct the scene details in the regions rather missing.
We detail other design parameters such as layer type, number of filters in the supplement. %

\vspace{-8 pt} %
\subsubsection{Mixer Network (Mix)}

The Mixer network is designed to augment the outputs ($I_i$) of the SRNet at different time locations ($i{\scriptstyle=}\{n{\scriptstyle-}m, n ,n{\scriptstyle+}m\}$) to reconstruct detail-rich intensity image ($O_n$) at the central stack's timestamp ($n$). 
This network employs convolutional layers to reconstruct the intensity image with fine details.

\vspace{-8 pt} %
\subsubsection{Similarity Loss (Sim)}
\label{criterion}

Given a reconstructed image ($O$) and its GT ($G$), we define a loss function with two terms.
First, we use an unstructured loss such as the $\ell_1$ norm to reconstruct overall sharper images as $\mathcal{L}_{\ell_1}(O,G) = \| O-G \|_1$ rather than $\ell_2$ which results in smoothed edges with low frequency texture in output images.
As the $\ell_1$ may lose the structural information of a scene, we further leverage a criterion capable of compensating the lack of structure by the Learned Perceptual Image Patch Similarity (LPIPS) or perceptual similarity \cite{zhang2018unreasonable} as the second term of our objective function.
Specifically, given a pair of images ($O,G$) encoded by a pretrained network (\eg, AlexNet \cite{krizhevsky2012imagenet}), the near end features ($\hat{G}^l_{hw}$) of the $l^\text{th}$ layer are extracted while its activations are normalized by the channel dimension ($H_l,W_l$). 
Then, each channel is scaled by a vector $w_l$ \cite{zhang2018unreasonable}, and the $\ell_2$ distance is computed.
Finally, a spatial mean is computed over the image axes ($h,w$) through all layers ($l$) for the LPIPS loss as %
\begin{equation}
\resizebox{.9\hsize}{!}{
$\mathcal{L}_{LPIPS}(O,G) = \sum_l \frac {1}{H_lW_l}\sum_{h,w}\| w_l \odot(\hat{O}^l_{hw} - \hat{G}^l_{hw}) \|_2^2.$
}
\label{eq:loss_PS}
\end{equation}
The final objective function, $\mathcal{L}_{sim}$, is the combination of the both terms with a balancing parameter $\lambda$ as 
\begin{equation}
\mathcal{L}_{sim}(O,G) = \mathcal{L}_{\ell_1}(O,G) + \lambda\mathcal{L}_{LPIPS}(O,G),
\label{eq:loss_l1+PS}
\end{equation}
which we minimize to learn the parameters. %

\begin{table*}[t!]
    \caption{Comparison to state-of-the-art intensity synthesis methods on real-world sequences \cite{mueggler2017event}. Our method outperforms the previous methods in all sequences in LPIPS, and on average in SSIM. The runner up method is underlined. We used the reported numbers in \cite{rebecq2019high} for HF \cite{scheerlinck2018continuous}, MR \cite{reinbacher2016real} and EV \cite{rebecq2019high} while evaluated the authors' reconstructed images for EG \cite{mostafavi2019event}.}
    \centering
    \resizebox{0.99\linewidth}{!}{
    \begin{tabular}{cccccc>{\columncolor[gray]{0.9}}cccccc>{\columncolor[gray]{0.9}}cccccc>{\columncolor[gray]{0.9}}c}
    \toprule
    \rowcolor{white}
    && \multicolumn{5}{c}{SSIM ($\uparrow$)} & & \multicolumn{5}{c}{MSE ($\downarrow$) } & & \multicolumn{5}{c}{LPIPS ($\downarrow$)}\\
    \cmidrule{3-7} \cmidrule{9-13} \cmidrule{15-19}
    
    Sequence && HF \cite{scheerlinck2018continuous} & MR\cite{reinbacher2016real} & EV\cite{rebecq2019high} & EG\cite{mostafavi2019event} & Ours && HF \cite{scheerlinck2018continuous} & MR\cite{reinbacher2016real} & EV\cite{rebecq2019high} & EG\cite{mostafavi2019event} & Ours && HF \cite{scheerlinck2018continuous} & MR\cite{reinbacher2016real} & EV\cite{rebecq2019high} & EG\cite{mostafavi2019event} & Ours \\ 
    \midrule
    \lmss{dynamic\underline{ }6dof}  && 0.39 & \bf{0.52} & 0.46 & \underline{0.48} & 0.44 && 0.10 & \underline{0.05} & 0.14 & \bf{0.03} & \underline{0.05} && 0.54 & 0.50  & 0.46 & \underline{0.45} & \bf{0.42} \\
    \lmss{boxes\underline{ }6dof}    && 0.49 & 0.45 & \bf{0.62} & 0.45 & \underline{0.61} && 0.08 & 0.10 & 0.04 & \underline{0.03} & \bf{0.02} && 0.50 & 0.53  & \underline{0.38} & 0.48 & \bf{0.32} \\
    \lmss{poster\underline{ }6dof}   && 0.49 & 0.54 & \underline{0.62} & 0.61 & \bf{0.63} && 0.07 & 0.05 & 0.06 & \bf{0.01} & \underline{0.02} && 0.45 & 0.52 & \underline{0.35} & 0.42 & \bf{0.29}\\
    \lmss{shapes\underline{ }6dof}   && 0.50 & 0.51 & \bf{0.80} & 0.56 & \underline{0.79} && 0.09 & 0.19 & 0.04 & \underline{0.03} & \bf{0.01} && 0.61 & 0.64 & \underline{0.47} & 0.51 & \bf{0.38} \\
    \lmss{office\underline{ }zigzag} && 0.38 & 0.45 & 0.54 & \underline{0.67} & \bf{0.68} && 0.09 & 0.09 & 0.03 & \underline{0.01} & \underline{0.01} && 0.54 & 0.50 & \underline{0.41} & 0.36 & \bf{0.29} \\
    \lmss{slider\underline{ }depth}  && 0.50 & 0.50 & \underline{0.58} & 0.54 & \underline{0.59} && 0.06 & 0.07 & 0.05 & \underline{0.02} & \underline{0.02} && 0.50 & 0.55 & 0.44 & \underline{0.42}  & \bf{0.34} \\
    \lmss{calibration}               && 0.48 & 0.54 & \underline{0.70} & 0.67 & \bf{0.71} && 0.09 & 0.07 & \underline{0.02} & \underline{0.01} & \underline{0.01} && 0.48 & 0.47 & \underline{0.36} & 0.42 &  \bf{0.24} \\ \midrule
    Average                         && 0.46 & 0.50 & \underline{0.62} & 0.57 & \bf{0.64} && 0.08 & 0.09 & 0.05 & \underline{0.02} & \underline{0.02} && 0.52 & 0.53 & \underline{0.41} & 0.43 & \bf{0.33} \\
    \bottomrule
    \end{tabular}
    }
    \label{tab:realworld}
    \vspace{-10 pt}
\end{table*}
\begin{table}[t!]
    \caption{\protect \centering Quantitative comparison of super-resolved intensity images from events directly (Ours) to events to intensity image synthesis (EV) combined with SISR \cite{dai2019second} and MISR\cite{haris2019recurrent} methods.}
    \centering
    \resizebox{0.99\linewidth}{!}{
        \begin{tabular}{ccccc}
        \toprule
        Method  & PSNR ($\uparrow$) & SSIM ($\uparrow$)  & MSE ($\downarrow$)  & LPIPS ($\downarrow$) \\\midrule
        EV + SISR  $2\times$ & 11.292 &	0.384 &	0.348 &	0.394 \\
        EV + MISR  $2\times$ & \underline{11.309} &	\underline{0.385} &	\underline{0.347} &	\underline{0.392}\\
        \rowcolor{Gray}
        Ours $2\times$ & \textbf{16.420}	& \textbf{0.600} &   \textbf{0.108}  & \textbf{0.172}\\ \midrule
        EV + SISR  $4\times$ & 11.168 &	\underline{0.396} &	0.089 &	0.543\\
        EV + MISR  $4\times$ & \underline{11.293} &	0.384 &	\underline{0.087} &	\underline{0.396}\\
        \rowcolor{Gray}
        Ours $4\times$ & \textbf{16.068}	& \textbf{0.560} & \textbf{0.028} & \textbf{0.253}\\ \bottomrule
        \end{tabular}
    }
    \vspace{-1em}
    \label{tab:E2I2SR}
\end{table}

\section{Experiments and Analyses}

For the empirical validation, we use generated sequences using the event camera simulator (ESIM) \cite{rebecq2018esim} and four challenging and diverse real-world public datasets.%
\cite{bardow2016simultaneous,mueggler2017event,scheerlinck2018continuous, zhu2018multivehicle}. 
We describe the details of our dataset in the supplement.
For the quantitative analyses, we use PSNR in dB (logarithmic scale), the structural similarity \cite{wang2003multiscale} (SSIM) as a fraction between zero (less similar) to one (fully similar), the mean squared error (MSE), and the perceptual similarity (LPIPS) as a metric to evaluate the similarity of the high level features in two images (lower the value, more the similarity).
For each experiment, we train our network on a cluster of $8$ Titan-Xp GPUs. 
Batch size is $8$ and initial learning rate is $0.01$ which is decayed by a factor of $10$ at every half of the remaining epochs of the given maximum number of epochs (\eg, 50 in our experiments).
We use $\lambda=0.01$ for all our experiments, otherwise mentioned.

\subsection{Comparison with State of the Arts}
We are the first to propose the task of direct reconstruction SR intensity image from events thus there are no directly comparable methods.
So, we first down-sample our outputs and compare to same-size intensity reconstruction methods to evaluate the quality of our reconstruction. %
Then we compare our method to the state-of-the-art intensity reconstruction methods combined with the state-of-the-art super-resolution (SR) methods.

\vspace{-1em}\paragraph{Image reconstruction without super-resolution.} 
We compare down-sampled outputs of our method to the state-of-the-art event to intensity image methods on seven challenging real-world sequences from the Event Camera dataset \cite{mueggler2017event}. 
For notation brevity, we abbreviate the high pass filter method \cite{scheerlinck2018continuous} as HF, manifold regularization \cite{reinbacher2016real} as MR, event to video generation \cite{rebecq2019high} as EV and event to intensity by conditional GANs as EG \cite{mostafavi2019event}.
Following the evaluation protocols in many real-world event datasets \cite{mueggler2017event,scheerlinck2018continuous, zhu2018multivehicle}, we consider APS frame as GT.
We follow the sequence split of \cite{rebecq2019high} and use the reported performance measures of HF, MR and EV. For EG, we used the authors' reconstructed images to evaluate the performance. 

As shown in Table \ref{tab:realworld}, our proposed method outperforms all other methods in LPIPS. 
It implies that the reconstructed intensity image is perceptually better than the previous methods. 
Our method also exhibits higher SSIM scores on multiple sequences and comparable MSE errors to EG. 
Similar to EV, we train the model only with the synthetic sequences and apply to real world sequences. In this challenging zero-shot data transfer setting without fine-tuning, our method outperforms other methods on real-world events.
Note that the two runner up methods in LPIPS (EV and EG) also use learning based framework.

\vspace{-1em}\paragraph{Super-resolved image reconstruction.}

We now combine state-of-the-art event to intensity reconstruction algorithms with state-of the-art SR methods and compare our method to them.
For the state-of-the-art event to intensity algorithm, we use EV\footnote{Publicly available at {\scriptsize \url{https://github.com/uzh-rpg/rpg\_e2vid}}.} since it is the runner up method that outperforms EG in SSIM and LPIPS in most of the sequences and on average (Table~\ref{tab:realworld}).
For super resolution algorithms, we use two recent super-resolution algorithms; one for SISR \cite{dai2019second} and another for MISR \cite{haris2019recurrent}.
As shown in Table \ref{tab:E2I2SR}, our method outperforms the state-of-the-art intensity reconstruction algorithms combined with the state-of-the-art SR algorithms in all metrics by large margins.
We use 30 sequences from our generated dataset by ESIM.

\begin{figure}[t!]
\centering
    \begin{tabularx}{\linewidth}{
    >{\centering}X
    >{\centering}X
    >{\centering}X
    >{\centering}X
    >{\centering\arraybackslash}X
    }
     \scriptsize Events & \scriptsize EV &  \scriptsize EV+SR $2\scriptstyle\times$ &  \scriptsize Ours $2\scriptstyle\times$ & \scriptsize APS\\
     \end{tabularx}
\vspace{-6 pt}
\includegraphics[width=1\linewidth]{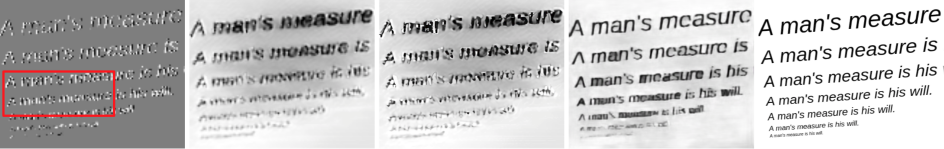}
    \vspace{-3 pt}
    \begin{tabularx}{\linewidth}{
    >{\centering}X
    >{\centering}X
    >{\centering}X
    >{\centering\arraybackslash}X
    }
    \scriptsize EV &  \scriptsize EV+SR $2\scriptstyle\times$  &  \scriptsize Ours $2\scriptstyle\times$ & \scriptsize APS\\
    \end{tabularx}
\vspace{-6 pt}
\includegraphics[width=1\linewidth]{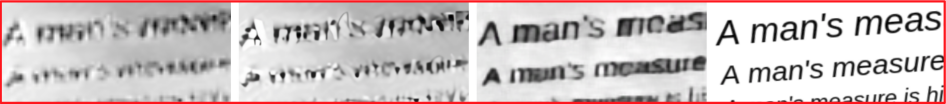}
    \vspace{-3 pt}
    \begin{tabularx}{\linewidth}{
    >{\centering}X
    >{\centering}X
    >{\centering}X
    >{\centering}X
    >{\centering\arraybackslash}X
    }
     \scriptsize Events & \scriptsize EV &  \scriptsize EV+SR $2\scriptstyle\times$ &  \scriptsize Ours $2\scriptstyle\times$ & \scriptsize APS\\
     \end{tabularx}
\vspace{-6 pt}
\includegraphics[width=1\linewidth]{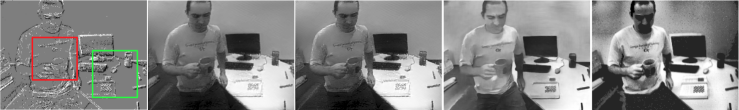}
    \vspace{-3 pt}
    \begin{tabularx}{\linewidth}{
    >{\centering}X
    >{\centering}X
    >{\centering}X
    >{\centering\arraybackslash}X
    }
    \scriptsize EV &  \scriptsize EV+SR $2\scriptstyle\times$  &  \scriptsize Ours $2\scriptstyle\times$ & \scriptsize APS\\
    \end{tabularx}
    \vspace{-6 pt}
    \includegraphics[width=1\linewidth]{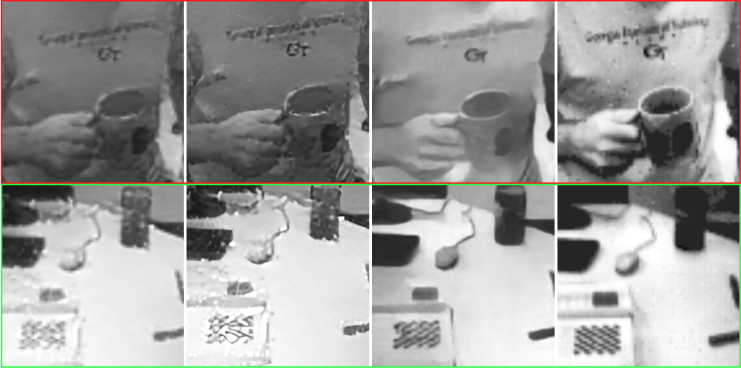}
    \vspace{-3 pt}
    \begin{tabularx}{\linewidth}{
    >{\centering}X
    >{\centering}X
    >{\centering}X
    >{\centering}X
    >{\centering\arraybackslash}X
    }
     \scriptsize Events & \scriptsize EV &  \scriptsize EV+SR $2\scriptstyle\times$ &  \scriptsize Ours $2\scriptstyle\times$ & \scriptsize APS\\
     \end{tabularx}
    \vspace{-6 pt}
    \includegraphics[width=1\linewidth]{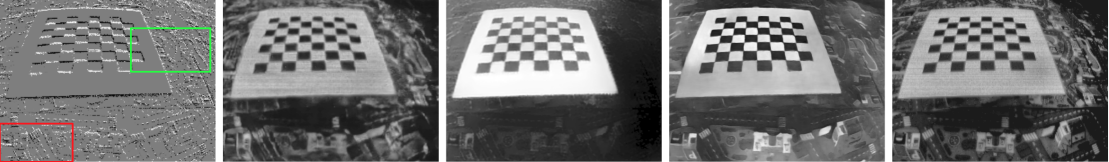}
    \vspace{-3 pt}
    \begin{tabularx}{\linewidth}{
    >{\centering}X
    >{\centering}X
    >{\centering}X
    >{\centering\arraybackslash}X
    }
    \scriptsize EV &  \scriptsize EV+SR $2\scriptstyle\times$  &  \scriptsize Ours $2\scriptstyle\times$ & \scriptsize APS\\
    \end{tabularx}
    \includegraphics[width=1\linewidth]{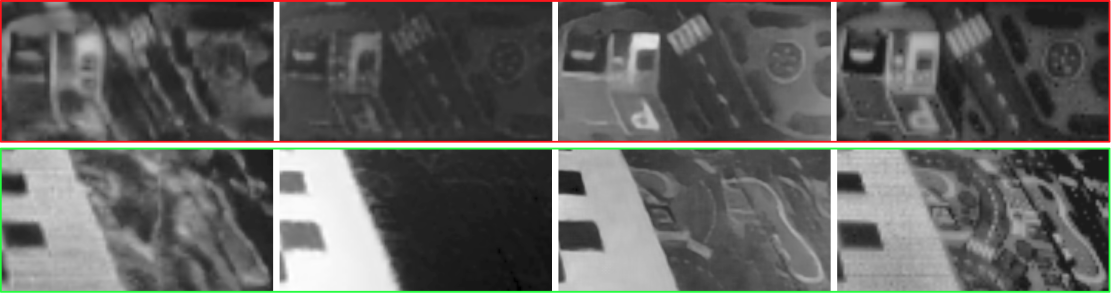}
    \vspace{-10 pt}
    \caption{Qualitative comparison among synthesizing SR intensity images directly (ours) and super-resolving as a downstream application to intensity image estimation (EV+MISR). %
    Highlighted boxes are zoomed for better comparison.}%
    \label{E2SR}
    \vspace{-1em}
\end{figure}

For qualitative analyses, we demonstrate intensity reconstruction by EV, the combination of EV+MISR and our method on real-world and simulated sequences in the Fig. \ref{E2SR} and Fig. \ref{teaser}. 
Note that our method reconstructs fine details from events.
In Fig. \ref{teaser}, EG does not always reconstruct scene details from the events and sometimes hallucinates jittery artifacts. 
While EV reconstructs scene details from the events relatively better than EG, it creates a shadow-like artifact and darkens some areas of the scene.
Furthermore, in the presence of hot pixels in the data, EV does not filter them; white or black dots appear in the results by EV while our method mostly filters them out without explicit operations to remove.
We present more results in the supplementary material.

\begin{figure*}[t!]
    \begin{tabularx}{\linewidth}{
    >{\centering}X
    >{\centering}X
    >{\centering}X
    >{\centering}X
    >{\centering}X
    >{\centering}X
    >{\centering}X
    >{\centering\arraybackslash}X
    }
    \scriptsize Events  & \scriptsize EG  &  \scriptsize EV  &  \scriptsize Ours & \scriptsize Events  & \scriptsize EG  &  \scriptsize EV  &  \scriptsize Ours\\
    \end{tabularx}
\includegraphics[width=1\linewidth]{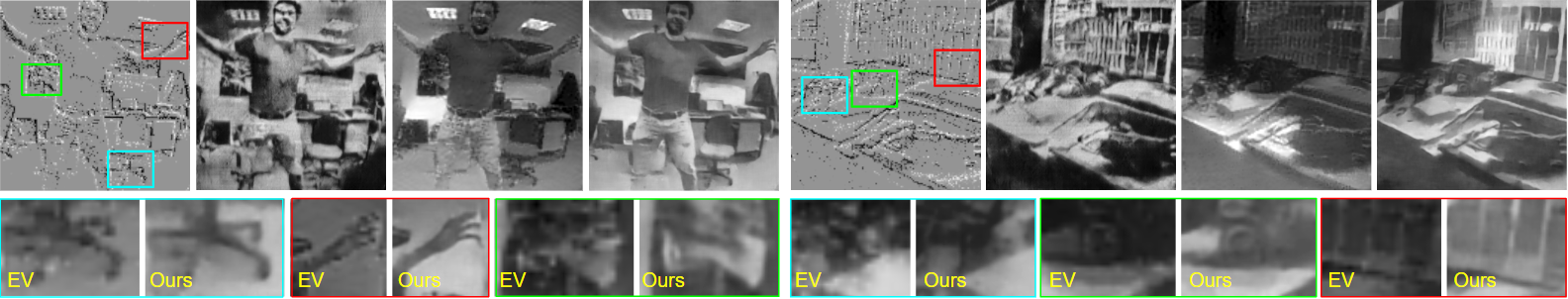}
\caption{Qualitative comparison of our downscaled outputs to EV ands EG on sequences from \cite{bardow2016simultaneous} (without APS). Our method is able to reconstruct structural details from inputs as small as $128{\scriptstyle\times}128$ pixels. More results are provided in the supplementary material.
}
\label{Bardow}
\vspace{-1 em}
\end{figure*}

We further conduct experiments on the sequences from another popular dataset \cite{bardow2016simultaneous} and qualitatively compare our method to EG and EV in Fig. \ref{Bardow}. 
Our method can reveal details that is not visible in constructing the same sized images such as fingertips or texture.

\subsection{Analysis on Loss Terms ($\mathcal{L}_{sim}$)}

We ablate the loss function to investigate the effect of each terms on image reconstruction quantitatively in Table \ref{tab:ablate} and qualitatively in Fig. \ref{l1_ps}. 
All analyses and ablation studies were performed with the simulated data for reliable quantitative analyses with high quality GT. 
Using only $\mathcal{L}_{\ell_1}$ term, we observe better performance in PSNR but leads to visually less sharp images thus low performance in all other metrics. Using only $\mathcal{L}_{LPIPS}$ term, we observe that images look visually acceptable but with the downside of lower PSNR with dot-like artifacts on regions with less events and on the edges.
The final proposed loss function $\mathcal{L}_{sim}$ performs the best in SSIM and MSE with a slight decrease in PSNR and LPIPS but creates visually the most plausible images.
\begin{table}[t!]
    \centering
    \caption{\protect \centering Ablation study of the loss function.} %
    \resizebox{0.95\linewidth}{!}{
    \begin{tabular}{ccccc}
    \toprule
    Loss  & PSNR ($\uparrow$) & SSIM ($\uparrow$)  & MSE ($\downarrow$)  & LPIPS ($\downarrow$) \\
    \midrule
    $\mathcal{L}_{\ell_1}$ & \textbf{15.33}	& \underline{0.517} & \underline{0.034} & 0.485 \\
    $\mathcal{L}_{LPIPS}$ & 10.06	& 0.388 & 0.454 & \textbf{0.232} \\
    \midrule
    $\mathcal{L}_{sim}$ (Full) & \underline{15.03}	& \textbf{0.528} &	\textbf{0.032}  & \underline{0.258} \\ 
    \bottomrule
    \end{tabular}
    }
    \label{tab:ablate}
    \begin{tabularx}{\linewidth}{
    >{\centering}X
    >{\centering}X
    >{\centering}X
    >{\centering}X
    >{\centering}X
    >{\centering\arraybackslash}X
    }
    $\scriptscriptstyle \ell_1$  & \tiny LPIPS  &  $\scriptscriptstyle\ell_1{+}$\tiny LPIPS  &  $\scriptscriptstyle\ell_1$  & \tiny LPIPS  & $\scriptscriptstyle\ell_1{+}$\tiny LPIPS\\
    \end{tabularx}   
\centering
        \includegraphics[width=1\linewidth]{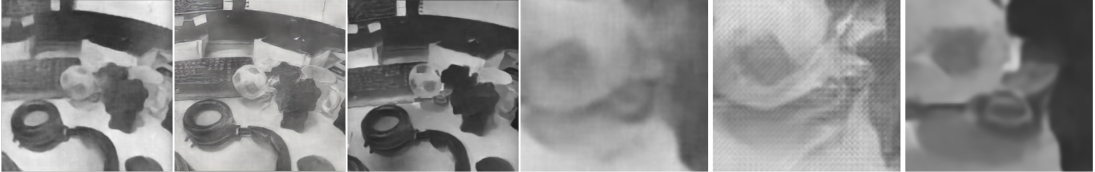}
\captionof{figure}{Effect of loss function on reconstruction quality. %
$\ell_1$ norm smooths edges, perceptual similarity (LPIPS) adds structural details but also creates artifacts. The combination of $\ell_1{\scriptstyle+}LPIPS$ ($\mathcal{L}_{sim}$) shows less artifacts while adding structural details.}
\label{l1_ps}
\vspace{0.5em}
    \caption{Effect of number of stacks and scale factor.}
    \centering
    \resizebox{0.99\linewidth}{!}{
    \begin{tabular}{cccccc}
    \toprule
    Scale & \# Stacks  & PSNR ($\uparrow$) & SSIM ($\uparrow$)  & MSE ($\downarrow$)  & LPIPS ($\downarrow$) \\\midrule
    \multirowcell{2}{$2\times$}
            & 3S & 15.46   & 0.554 &	0.323  & 0.191\\
            & 7S & \textbf{16.42}	& \textbf{0.600} &   \textbf{0.108}  & \textbf{0.172}\\ \midrule
    \multirowcell{2}{$4\times$} 
            & 3S & 15.03	& 0.528 &	0.032  & 0.258\\
            & 7S & \textbf{16.06}	& \textbf{0.560} &   \textbf{0.028}  & \textbf{0.253}\\ 
    \bottomrule
    \end{tabular}}
    \label{tab:scale_frame}

\vspace{-1em}
\end{table}

\subsection{Analysis on Super Resolution Parameters}

We evaluate the effect of two SR parameters; the upscale factor ($2\times, 4\times$) and size of the sequence of stacks ($3S,7S$) on the output quality.
We summarize the results in Table \ref{tab:scale_frame}.
Comparing $3S$ and $7S$, we observe that $7S$ results in better performance in all metrics. %
It implies that a longer recursion on the sequences may produce more reliable hidden states and results in better quality output. 
Also, when using longer sequences, it is more likely to capture events that happen only for a short period of time since unrolling on a larger recursion helps to keep information of short events.
It is more challenging to super resolve events to larger images as it is not trivial for an algorithm to handle large spatial locations where no events exist. 
Although the MSE has decreased, compared to $2\times$, it is because the number in the denominator is larger due to the size of the image and not much related to the output quality.

\subsection{Qualitative Analysis on HDR Sequences} %

\begin{figure}[t!]
    \begin{tabularx}{\linewidth}{
    >{\centering}X
    >{\centering}X
    >{\centering}X
    >{\centering\arraybackslash}X
    }
    \scriptsize Events \hspace{20 pt} & \scriptsize EV \hspace{20 pt} &  \scriptsize Ours $2\scriptstyle\times$ \hspace{20 pt} &  \scriptsize APS \\
    \end{tabularx}
        \includegraphics[width=1\linewidth]{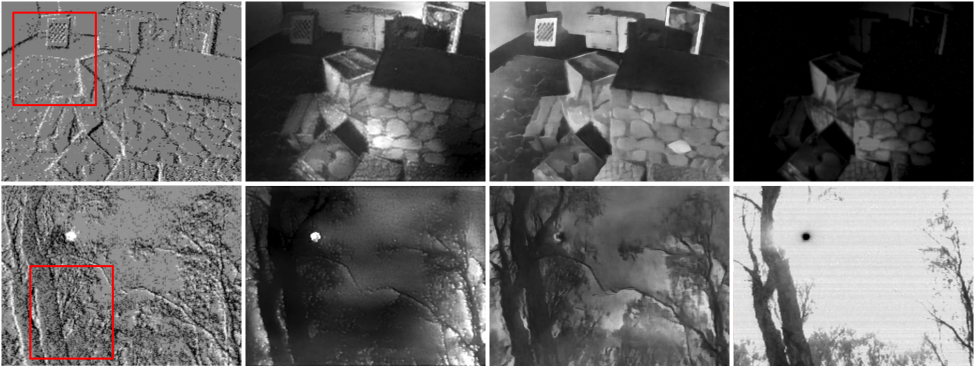}
\captionof{figure}{Image reconstruction comparison in extreme HDR scenarios \cite{mueggler2017event,scheerlinck2018continuous}. Our method synthesizes more details while producing less artifacts compared to EV and the APS. Please zoom in and compare the suggested red boxes.}
\label{fig:boxes_sun}
\vspace{-1em}
 \end{figure}

One challenging scenario using the event camera is to capture events under extreme dynamic range. We qualitatively analyze outputs under such extreme conditions and compare them to EV in Fig. \ref{fig:boxes_sun}. Normal cameras including the APS frame have much lower dynamic range and either create black regions (when the camera misses to sense intensity details under its sensing range as shown in the top row) or white regions (when light floods in the camera and the camera cannot sense higher than its sensing range as shown in the bottom row).  We observe that our method can address a higher range and reveal more structural details that EV and the APS frame fail to capture.

\subsection{Analysis on the Failure Modes}

Failure cases are mostly related to missing background details over long trajectories when the foreground objects have rapid movements. In such sequences, our method only recovers parts of the scene that are in a limited temporal distance to our central stack. 
We showcase and further analyze a number of failure modes in the supplementary material.

\section{Extensions}

\paragraph{Video reconstruction.}
We aim to reconstruct a single image not a video. 
So, the temporal consistency between frames are out of our interest thus not always held.
To extend our method to video reconstruction, we utilize a blind post-processing method \cite{Lai-ECCV-2018} to encode temporal consistency among the intensity images and demonstrate the qualitative results in a supplementary video.
To quantitatively evaluate the temporal consistency, we follow the temporal stability metric from \cite{lai2018learning}, which is based on the flow warping error between two consecutive synthesized frames $(F_t,F_{t+1})$:
\begin{equation}
\resizebox{.9\hsize}{!}{
    $E_{warp}(F_t,F_{t+1})= \frac{1}{\sum_{i=1}^{N}M_t ^{(i)}}\sum_{i=1}^{N}M_t ^{(i)} || F_t ^{(i)} - \hat{F}_{t+1} ^{(i)} ||_2^2, $
    }
\label{eq:warp}
\vspace{-5pt}
\end{equation}
where $\hat{F}_{t+1}$ is the warped frame of $F_{t+1}$ and $M_t \in \{0,1\}$ is the non-occlusion mask based on \cite{ruder2016artistic} to ensure that the calculations are applied only in the non-occluded regions.
We compute the optical flow used for warping the frame and the non-occlusion map based on the APS frames for evaluating the warping errors of all methods compared and APS as they are the GT. %
We summarize the results with different size of sequences ($3S$ and $7S$) in comparison to EV in Table \ref{tab:temp_consist}. 
While our methods ($3S$ and $7S$) are worse than EV due to lack of temporal consistency, a simple post-processing ($3S+$ and $7S+$) significantly improves the performance, outperforming both the EV \cite{rebecq2019high} and its post-processed version ($EV+$) by large margins.

\begin{table}[t!]
    \centering
    \captionof{table}{Temporal stability error evaluation (Eq. \ref{eq:warp}). Plus sign indicates blind post-processing \cite{lai2018learning}. Our method ($3S$, $7S$) does not directly consider temporal consistency however longer sequences of stacks ($7S$) are more consistent. EV\cite{rebecq2019high} uses up to $L{=}40$ input stacks and is initially more consistent. However, we get lower errors even on our smallest sequence after post-processing.}
    \resizebox{0.99\linewidth}{!}{%
    \begin{tabular}{cccccccc}
    \toprule
    $E_{warp}(\downarrow)$ & APS & $3S$ & $7S$ & EV \cite{rebecq2019high} & $3S+$ & $7S+$ & EV \cite{rebecq2019high}$+$\\
    \cmidrule(lr){1-1} \cmidrule(lr){2-5} \cmidrule(lr){6-8}
    \lmss{dynamic\underline{ }6dof}  & 0.61 & 20.35 & \underline{16.54} & \textbf{8.78}  & \textbf{3.42} & \underline{3.71} & 5.56\\
    \lmss{boxes\underline{ }6dof}    & 1.81 & \underline{16.69} & 17.51 & \textbf{15.69} & \textbf{3.58} & \underline{3.95} & 9.36 \\
    \lmss{poster\underline{ }6dof}   & 1.10 & \underline{18.80} & 22.66 & \textbf{17.74} & \textbf{4.41} & 5.91 & \underline{5.56} \\
    \lmss{shapes\underline{ }6dof}   & 0.44 & 24.00 & \underline{21.23} & \textbf{16.66} & \underline{2.80} & \textbf{2.63} & 8.33 \\
    \lmss{office\underline{ }zigzag} & 0.08 & 3.62 & \underline{2.19} & \textbf{0.72}    & \underline{0.36} & \textbf{0.34} & 0.44\\
    \lmss{slider\underline{ }depth}  & 0.02 & 0.57 & \underline{0.34} & \textbf{0.19}    & \underline{0.06} & \textbf{0.04} & 0.12\\
    \lmss{calibration}               & 0.36 & 15.46 & \underline{9.72} & \textbf{2.99}   & \textbf{1.31} & \underline{1.24} & 1.62\\ 
    \cmidrule(lr){1-1} \cmidrule(lr){2-5} \cmidrule(lr){6-8}
    Average                          & 0.63 & 14.21 & \underline{12.89} & \textbf{8.97}  & \textbf{2.28} & \underline{2.55} &  5.20\\ 
    \bottomrule
    \end{tabular}
    }
    \label{tab:temp_consist}
    \vspace{-1em}
 \end{table}

\vspace{-1em}\paragraph{Complementary and Duo-Pass.}
To evaluate our method in a challenging set-up, we do not use APS frame to super resolve images. Using APS frame, we can further improve quality of output. %
We name the extension by using APS frame as \emph{Complementary} \cite{scheerlinck2018continuous} or \emph{Comp.}
We train the initial state of network with the low resolution (LR) APS frame as a central stack (Sec. \ref{sec:overall_structure}) and provide events as its nearby stacks.
We observe that the network learns to add higher resolution details from the LR input. 

However, the Complementary method is sensitive to the quality of central stack, specifically if it is blurry or noisy, its artifacts are propagated to the final reconstruction.
To avoid such shortcoming, we propose another extension that does not use APS frames but use two iterations or passes from events only, called \emph{Duo-Pass}. 
In the first pass, we use the main scheme to create intensity images from events only. 
In the second pass, we use the synthesized intensity image from the first pass as the central stack similar to that we use the APS frame in the Complementary method.
By the Duo-Pass, we are able to further recover HR details that the first pass misses without the help of the APS frame. %
We qualitatively compare the results by our method (main), by the Duo-Pass and by the Comp. in Fig. \ref{fig:duo_comp}.
We provide more results in the supplementary material.

 \begin{figure}
     \centering
    \begin{tabularx}{\linewidth}{
    >{\centering}X
    >{\centering}X
    >{\centering}X
    >{\centering\arraybackslash}X
    }
    \scriptsize Ours (main) & \scriptsize Ours (Duo-Pass) &  \scriptsize APS  &  \scriptsize Ours (Comp.) \\
    \end{tabularx}     
        \includegraphics[width=1\linewidth]{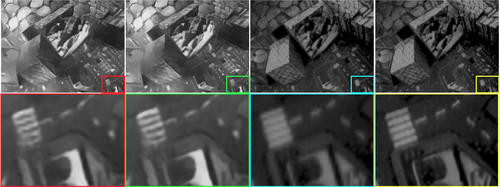}
\captionof{figure}{Extensions. Duo-Pass that iterates the SR twice and Complementary (Comp.) that uses events with APS frames.}
\label{fig:duo_comp}
\vspace{-1em}
 \end{figure}

\section{Conclusion}
We propose to directly reconstruct higher resolution intensity images from events by an end-to-end neural network. 
We demonstrate that our method reconstructs high quality images with fine details in comparison to the state of the arts in both the same size image reconstruction and super-resolution.
We further extend our method to the \emph{Duo-Pass} which performs an extra pass to add missing details and the \emph{Complementary} that utilizes APS frames in addition to events.
We also reconstruct videos by our method with a simple post-processing to ensure temporal consistency.

\vspace{1em}
\begingroup
{
\small
\noindent
\textbf{Acknowledgement.} This work was partly supported by the National Research Foundation of Korea (NRF) grant funded by the Korea government (MSIT) (No.2019R1C1C1009283) and (NRF-2018R1A2B3008640), the Next-Generation Information Computing Development Program through the NRF funded by MSIT, ICT (NRF-2017M3C4A7069369), Institute of Information \& communications Technology Planning \& Evaluation (IITP) grant funded by the Korea government (MSIT) (No.2019-0-01842, Artificial Intelligence Graduate School Program (GIST)) and (No.2019-0-01351, Development of Ultra Low-Power Mobile Deep Learning Semiconductor With Compression/Decompression of Activation/Kernel Data).\par
}
\endgroup
\newpage
{\small
\bibliographystyle{ieee_fullname}
\bibliography{egbib}
}
\newpage

\appendix 

\section*{Appendix}

\section{Discussions on Overlapped Stacking} %
\begin{figure}[b!]
\centering
        \includegraphics[width=1\linewidth]{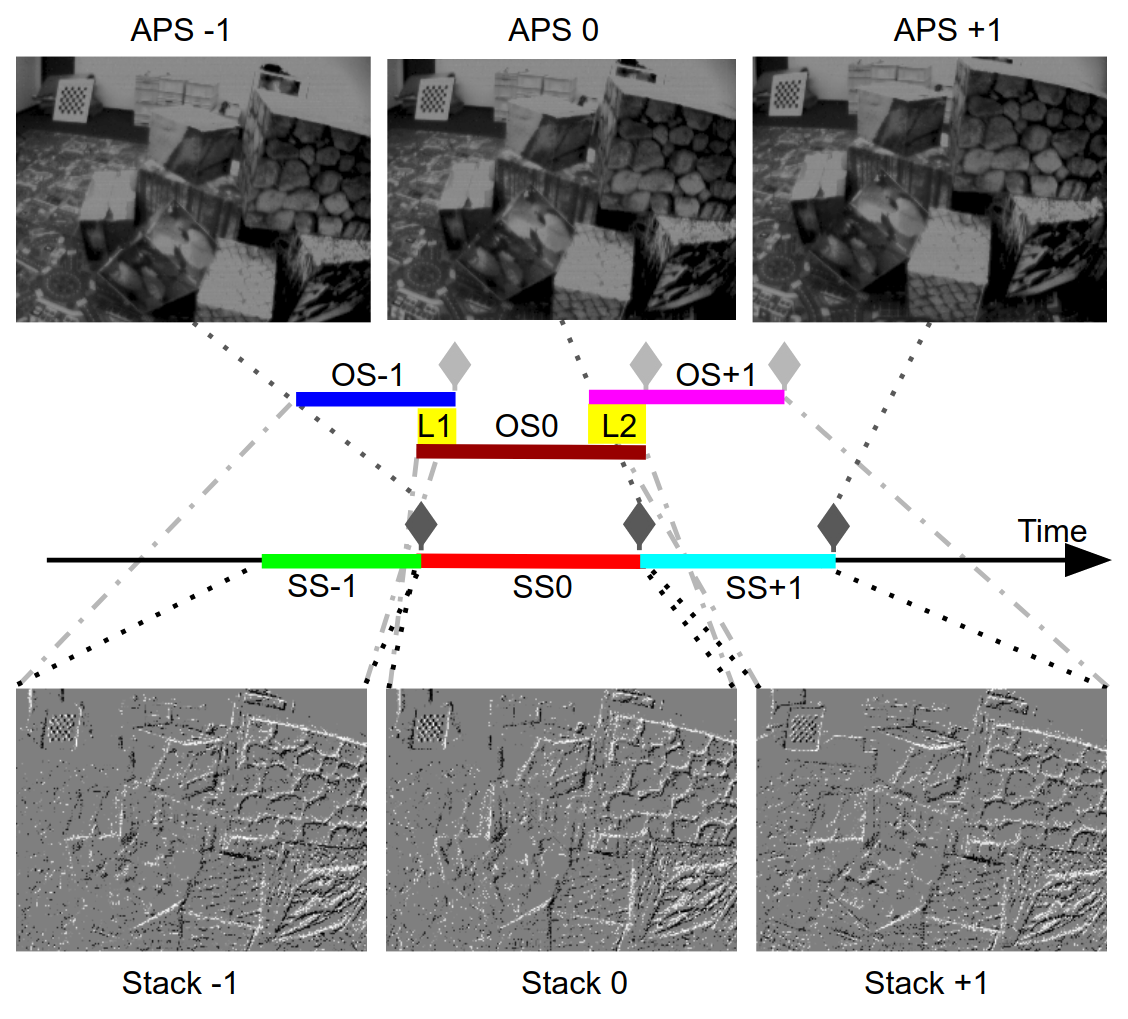}
\caption{Stacking with separate stacks (SS) or with overlapping Stacks (OS) in a sequence of $3S$. APS frame locations are shown as dark gray diamonds. Light gray diamonds show the location of virtual APS frames which are used in testing and do not respond to an actual APS frame. Central stack is shown as $Stack 0$ and the next ($+$) or previous ($-$) stacks with regards to the central stack are also shown. The yellow part shows the amount of shared overlapping events ($L1$ and $L2$). Note that the amount of events sets the length of the stack in time which will not necessarily be the same from one stack or overlapping region to another.}
\label{fig:stack}
\end{figure}

Based on the representation of the event stream stacked over time in Fig. 3 of the main paper, we are able to change the amount of overlap for stacking. This is demonstrated in Fig. \ref{fig:stack} where the location of APS frames are shown and events cover different amount of the stream over time as stacks based on how fast the events are fired which is related to the camera or scene speed movement. This means that the size of the colored stacks or the overlaps are not necessarily equal to each other. Two stacks can have common events up to a single event but less common events are desired to produce meaningful different images.

Unlike stacking based on time (SBT), stacking based on number of events (SBN) can consume different amount of time per stack which is related to the amount of events triggered from the scene. Furthermore, a stack might even surpass the location of the previous or next APS frame location and is not bound to the APS. This overlap is useful when there is large amount of scene movement and can prevent short-time fired events from being less effective by having them in more than one stack over the total number of stacks.

\section{Design Parameters for $SRNet$} %

\begin{figure*}[t!]
\centering
\includegraphics[width=1\linewidth]{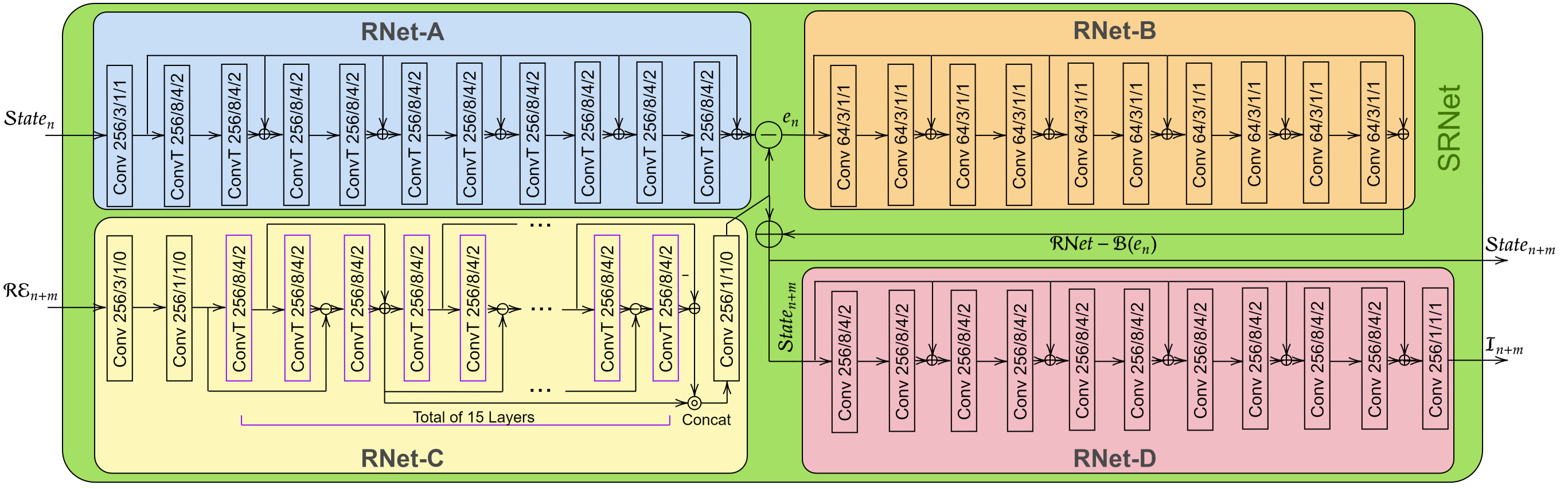}
\caption{SRNet in detail components. Colors following Fig. 4 of main paper.}
\label{fig:arch_design}
\vspace{-5 pt}
\end{figure*}

We illustrate the detailed design of our main super resolution network, the $SRNet$, in Fig.~\ref{fig:arch_design}.
The text in each box indicates layer type, number of filters, kernel size, stride and padding respectively (\eg, Conv 64/3/1/1).
The projection-wise setting of the recurrent residual modules follows the well-known iterative procedure for super-resolving multiple LR features called back-projection \cite{irani1990super}. 
We adopt the idea to design our $SRNet$; more specifically $RNet{\text -}B$ performs back-projection from $RE_{m+n}$ to $State_n$ for producing the residual $RNetB(e_n)$.

\section{The Synthetic Dataset} %
\label{sec:dataste_creation}

\subsection{Background}

We create a dataset using the event camera simulator (ESIM) \cite{rebecq2018esim} for high quality GT as many real world datasets have following issues, making the evaluation less reliable.

\vspace{-1em}\paragraph{Imperfect APS frame as groundtruth (GT).} 
The event camera needs movements of the scene or the camera to produce outputs but
rapid movements create motion blur on the intensity image. 
In addition, the dynamic range of an event camera and an intensity camera are much different which one device might sense parts of the scene that the other device does not. The combination of these factors makes real-world sensing devices prone to errors when used as the training source.

\vspace{-1em}\paragraph{Lack of high resolution GT.} When working with the same resolution intensity and event device for creating low resolution (LR) event and high resolution (HR) intensity pairs, one might think of resizing the event data to a smaller value. This might work on the training set but when generalizing to the test set that does not have such resized inputs such as the original events of the event camera, the outputs will have much lower quality. The reason behind it is that subsampling algorithms leave unwanted traces on the event stack. This artifacts might seem negligible, but in a learning based solution, it leads to learning erroneous parameters. In our experiments, subsampling the events resulted in a drop of almost $2 dB$ in terms of PSNR. This is a crucial step to reach higher quality outputs for cross evaluating on other datasets which we set as a goal.

As a remedy, we utilize a pair of synthetic cameras with different resolutions. We set the LR (event) camera has $128{\scriptstyle\times}128$ pixels resolution and the HR intensity camera has $256{\scriptstyle\times}256$ or $512{\scriptstyle\times}512$ pixels based on the upscale factor ($2\times$ or $4\times$), both sharing the same camera center. To have exactly the same filed of view in both cameras without further warping requirements, the focal length is multiplied to the desired upscale factor when moving from the LR event camera to the HR GT intensity camera.

\subsection{Dataset Detail}

We created our dataset using $1,000$ different images from the Microsoft COCO 2017 unlabeled images \cite{lin2014microsoft} placed on a planar surface while moving the cameras in 6-DoF on top using random trajectories and created almost $120K$ sequences of stacks. %
Different cameras can have different threshold values, therefore we randomly set the positive and negative threshold independently for each sequence to prevent the network from adapting to this parameter therefore being versatile to the input source all following the implementation details of \cite{rebecq2019high}.
Although we train our network only with the simulated dataset, we can fully transfer to real-world scenes without any fine tuning in a complete blind dataset transfer setting.

\section{Additional Qualitative Results} %
\paragraph{Comparison to the State of The Arts.}
We present more results on real-world and simulated sequences in Fig. \ref{fig:bardow_b}, \ref{fig:clock} and \ref{fig:outdoor}.

\vspace{-1em}\paragraph{Results on dataset \cite{rebecq2019high}.} %
Furthermore, we used the new dataset in \cite{rebecq2019high} with includes challenging sequences with high dynamic range and in high-speed scenarios. 
We showcase a sample in the high speed scenario of popping a water balloon over time in Fig. \ref{fig:popping_water_balloon}. 
Our method is able to reconstruct super-resolved details from the background scene and the fast moving foreground objects.

\section{Failure Mode Anlaysis} %
Since the largest number of stacks in a sequence we use was 7 (in $S7$), we are not able to recover the farther events over the 7 stacks due to limited GPU resources. Therefore, our algorithm may miss some background detail when fast foreground moving objects fire large amount of events that make the stacking exceed the 7 stacks. 
Fig \ref{fig:limitation} demonstrates a sample condition shown in a sequential manner over time.

Furthermore, if the events in a stream are noisy or dead pixels exist our method will create blurry artifacts in the presentation of those events. Parts of the stream used in Fig. \ref{fig:failure1} suffer from blurry artifacts. 
The final reconstruction artifacts are attributed to the lack of events when the camera movement is parallel to the scene structure, therefore events will not fire as shown in Fig. \ref{fig:failure2}. This artifact is often found in many reconstruction methods based on pure events.

\begin{figure*}[t!]
\centering
        \includegraphics[width=1\linewidth]{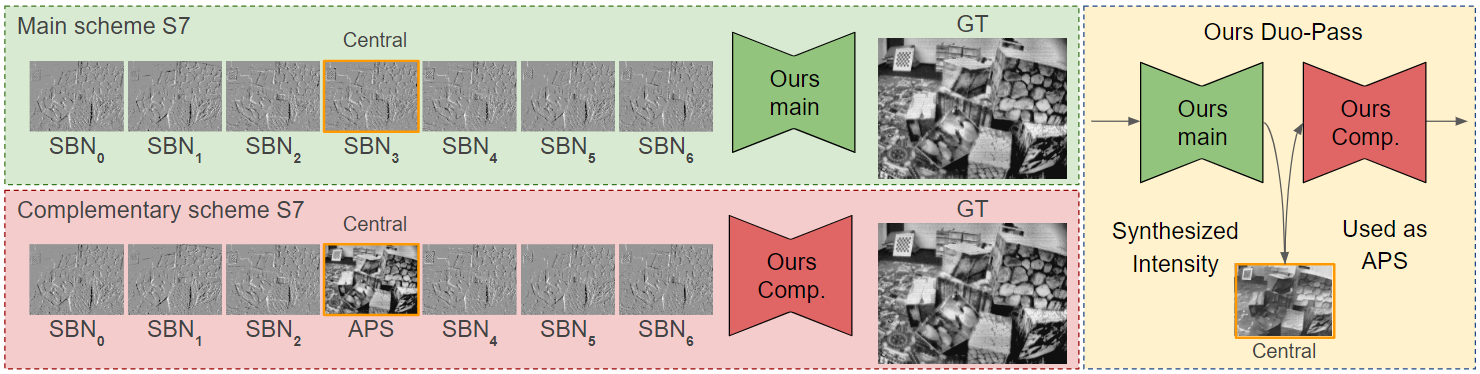}
\caption{Main and complimentary (Comp.) scheme with $S7$ stacks in a sequence. The central stack is highlighted in the middle and all other stacks will be compared to this stack fore optical flow creation. By putting the main network's output from pure events as a LR input (central stack) for the Comp. network we can have the Duo-Pass network which can add more details to the original intensity image.}
\label{fig:orig_com_duo}
\vspace{-5 pt}
\end{figure*}
\section{Details of Extensions with More Results} %
\subsection{Complimentary}
The Complementary extension uses the available APS frame and the events together to make a higher resolution intensity image by fusing the best from both sources. The training process, described in the main manuscript, is shown for seven stacks in a sequence ($S7$) in the green section of Fig. \ref{fig:orig_com_duo}. 
The central stack is highlighted in the middle ($SBN_3$). The complimentary training also follows the same process but instead of events as a central stack it has the low resolution version of the GT. Each previous or next stack will be fused with the LR GT (APS) and fed to the network. At inference, each event frame will add further HR details to the LR APS frame creating a super-resolved high quality output. Further results are shown in Fig. \ref{fig:OCD} and Fig. \ref{fig:OCD2}. 
There are two downsides of the Complementary approach; (1) we can only reach up to the frame rate of the APS since we use the APS frames, (2) if the LR APS is noisy and blurry, this artifacts will be propagated to the output image.

\subsection{Duo-Pass}
To avoid the downsides of the Complementary but obtain better quality output, 
we utilize the output of our method as the LR images in the complimentary extension, we can solve the noise and blur propagation and remove the frame-rate limitation. As shown in Fig. \ref{fig:orig_com_duo}, we just place the central stack (or frame) of the complementary method with the synthesised intensity image of our main method. We call this method a Duo-Pass and compare with Complementary and our original method in Fig. \ref{fig:OCD}.

\section{Additional Analysis on the Effect of Number of Events Per Stack}
The number of events in each stack affects the output reconstruction as shown in Fig. \ref{fig:1_5_10}. 
When the number of event are around $5,000$ events for image sizes of $240\scriptsize{\times}180$, the output is generally in a reasonable quality. %
However, adding much more events creates shadow-like outputs or blurred regions. Having much less events results in faded regions due to lack of information. Depending on the scene complexity, more or less events will be required for the best quality result.

To prevent overridden events in cases that the shapes on bottom last row, we stop adding events to the stack if a specific pixel gets overwritten more than $50$ times and continue with the next stack in the sequence. This is the general process while hand-tuning this number might get better results for specific cases.

These examples further show that the APS frame is not a good reference for comparing the reconstruction of events in terms of low dynamic range, motion blur and locations where events exist and there is no intensity details corresponding to it (\eg, in Fig 1 of the main manuscript under the table in the 3rd row) or locations where image details exist but no events have fired (tape and paper detailed areas around the shapes in the last row).

\section{Additional Analysis on the Effect of using Optical Flow by $FNet$}
Stacking events by definition causes loss of temporal relations among events. 
To recover that loss, we utilize FNet in our design by employing optical flow by following recent MISR techniques for inter-relating images over a sequences \cite{sajjadi2018frame,haris2019recurrent}.
In order to show the usefulness of optical flow on our intensity reconstruction, we ablate its effect by removing it and summarize the results in Table \ref{tab:ablate_flow}. The base network is design for $4\times$ scale and $3S$ stacks with $\ell_1$ norm only as the optimization criterion.
As shown in the table, without $FNet$ the performances are noticeably decreased in all metrics.

\begin{table}[t!]
    \caption{Ablating the existence of FNet. Adding FNet to $\ell_1$ improves all metrics. In the main paper all experiments included FNet and all ablations where performed using $4\times$ scale and 3 stacks ($3S$).}
    \vspace{.5em}
    \centering
    \resizebox{0.99\linewidth}{!}{
    \begin{tabular}{ccccc}
     \toprule
    Similarity  & PSNR ($\uparrow$) & SSIM ($\uparrow$)  & MSE ($\downarrow$)  & LPIPS ($\downarrow$) \\\midrule
    without FNet  & \underline{14.97}&\underline{0.505} & \underline{0.036} & \underline{0.499} \\
    with FNet & \textbf{15.33}	& \textbf{0.517} & \textbf{0.034} & \textbf{0.485} \\
    \bottomrule
    \end{tabular}}
    \label{tab:ablate_flow}
    \vspace{-1em}
\end{table}

\begin{figure}[t!]
    \begin{tabularx}{\linewidth}{
    >{\centering}X
    >{\centering}X
    >{\centering\arraybackslash}X
    }
    \scriptsize Events  & \scriptsize Reconstruction  &  \scriptsize APS\\
    \end{tabularx}
        \includegraphics[width=1\linewidth]{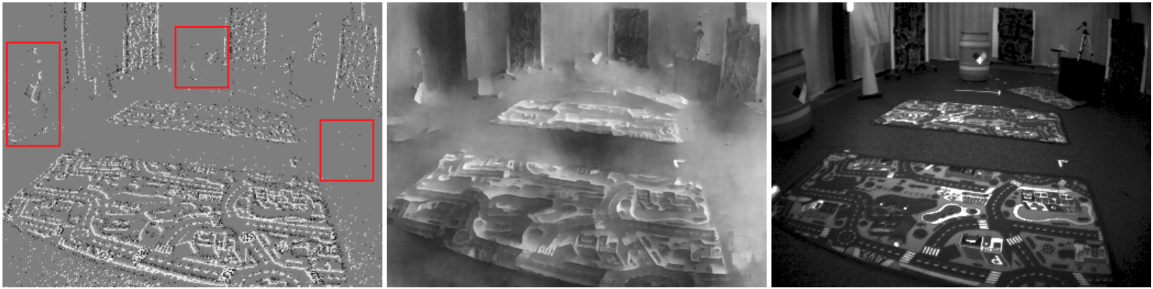}
\caption{Foggy edges in the presence of noisy events.}
\label{fig:failure1}
    \begin{tabularx}{\linewidth}{
    >{\centering}X
    >{\centering}X
    >{\centering\arraybackslash}X
    }
    \scriptsize Events  & \scriptsize Reconstruction  &  \scriptsize APS\\
    \end{tabularx}
        \includegraphics[width=1\linewidth]{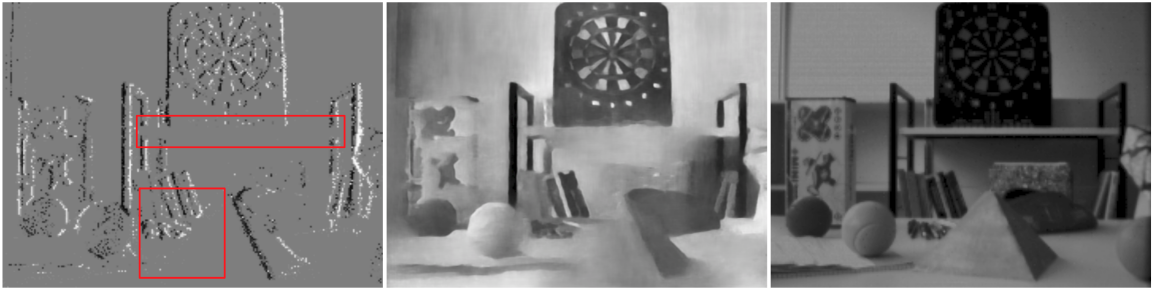}
\caption{Blur artifact due to lack of events.}
\label{fig:failure2}
\vspace{-6 pt}
\end{figure}

\section{Computational Complexity in Time}
The average run-time to super resolve an image with the scaling factor of ($2\times$, $4\times$) from the input event stacks with the dimension of $180{\times}240{\times}3$ holding 5,000 events per stack on a single Titan-Xp GPU, is%
: \{$(3S, 2{\times})$,$(3S, 4{\times})$,$(7S, 2{\times})$,$(7S, 4{\times})$\}$\rightarrow$\{18.5, 19.4, 250.8, 450.9\} (ms) 
where $3S$ and $7S$ refer to the number of stacks (3 and 7, respectively) in each sequence.

\begin{figure*}[t]
    \begin{tabularx}{\linewidth}{
    >{\centering}X
    >{\centering}X
    >{\centering}X
    >{\centering\arraybackslash}X
    }
    \scriptsize Ours Main  & \scriptsize Ours Duo-Pass  &  \scriptsize APS  &  \scriptsize Ours Comp.\\
    \end{tabularx}
        \includegraphics[width=1\linewidth]{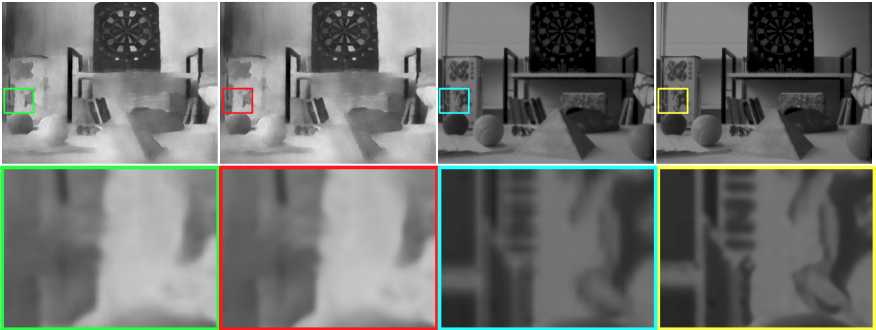}
        \includegraphics[width=1\linewidth]{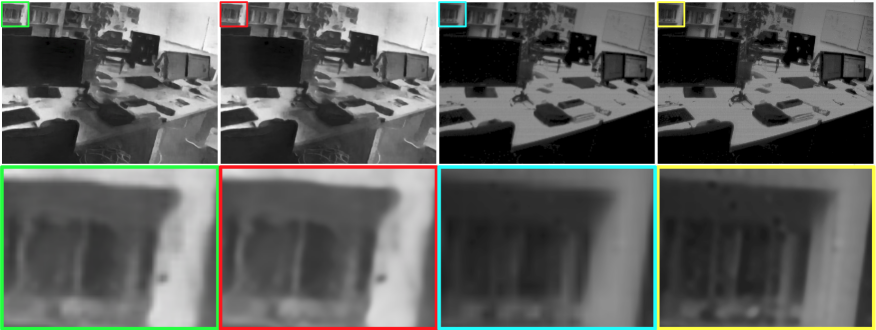}
        \includegraphics[width=1\linewidth]{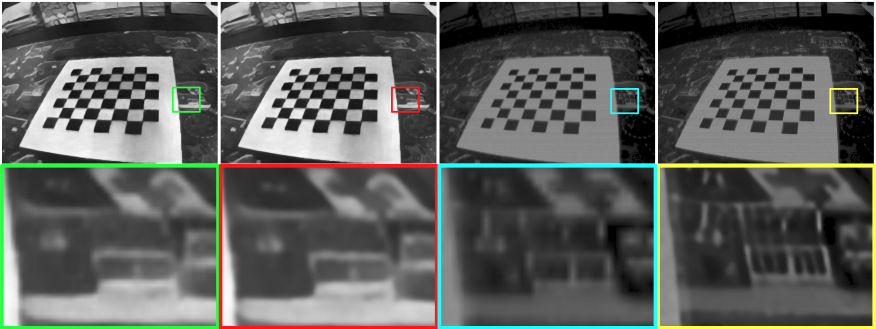}
\caption{More results of our main and extension methods of double passing (Duo-pass) and complementary processing (Comp.) on real-world dataset \cite{mueggler2017event}. Regions in the colored boxes are zoomed $20\times$ for comparison. APS frames that where very dark are histogram equalized for visualization only. High quality outputs can be achieved when complementing APS frames and events.}
\label{fig:OCD}
\end{figure*}

\begin{figure*}[t]
    \begin{tabularx}{\linewidth}{
    >{\centering}X
    >{\centering}X
    >{\centering}X
    >{\centering\arraybackslash}X
    }
    \scriptsize Ours Main  & \scriptsize Ours Duo-Pass  &  \scriptsize APS  &  \scriptsize Ours Comp.\\
    \end{tabularx}
        \includegraphics[width=1\linewidth]{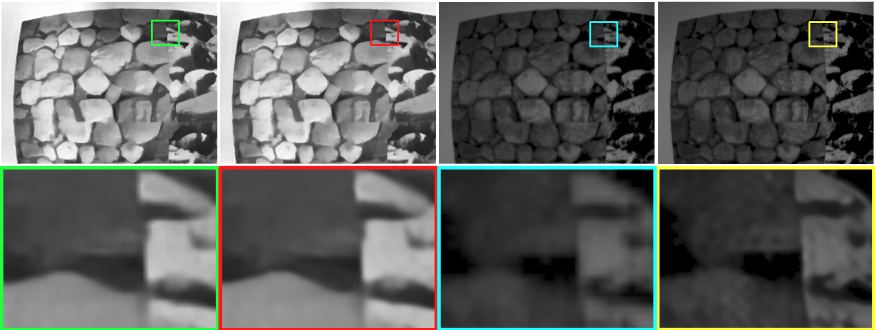}
        \includegraphics[width=1\linewidth]{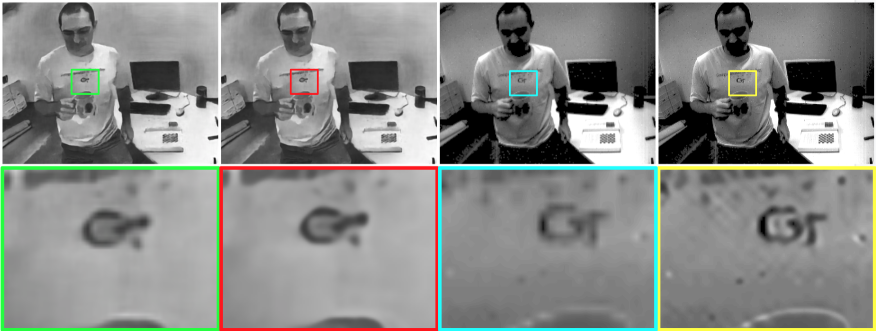}
        \includegraphics[width=1\linewidth]{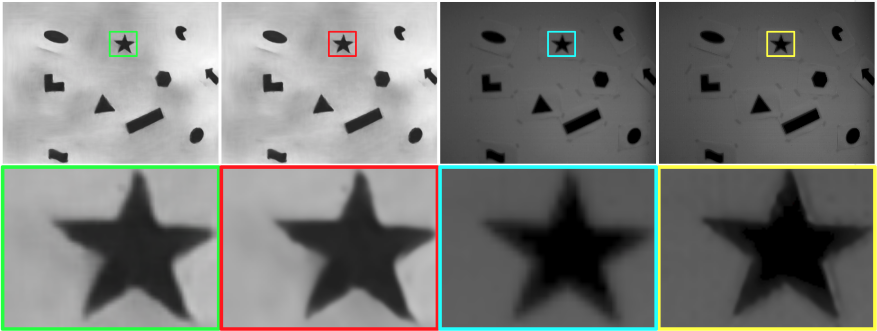}
\caption{More results of our main and extension methods of double passing (Duo-pass) and complementary processing (Comp.) on real-world dataset \cite{mueggler2017event}. Regions in the colored boxes are zoomed $20\times$ for comparison. APS frames that where very dark are histogram equalized for visualization only. High quality outputs can be achieved when complementing APS frames and events.}
\label{fig:OCD2}
\end{figure*}

\begin{figure*}[t!]
    \vspace{-1em}
    \begin{tabularx}{\linewidth}{
    >{\centering}X
    >{\centering}X
    >{\centering}X
    >{\centering}X
    >{\centering}X
    >{\centering}X
    >{\centering}X
    >{\centering\arraybackslash}X
    }
    \scriptsize Events  & \scriptsize EG  &  \scriptsize EV  &  \scriptsize Ours & \scriptsize Events  & \scriptsize EG  &  \scriptsize EV  &  \scriptsize Ours\\
    \end{tabularx}
        \includegraphics[width=\linewidth]{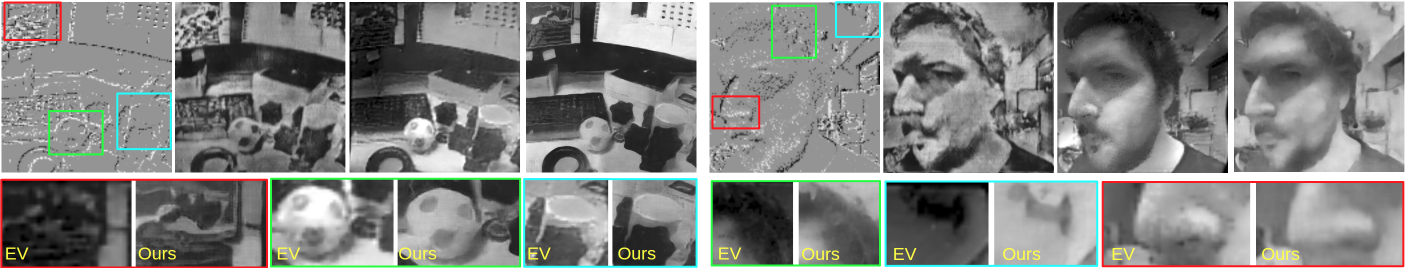}
\caption{Additional comparison between EV, EG and our results on sequences from \cite{bardow2016simultaneous} (In addition to the Fig. 5 of main manuscript).}
\label{fig:bardow_b}
\end{figure*}

\begin{figure*}[h!]
\centering
    \includegraphics[width=1\linewidth]{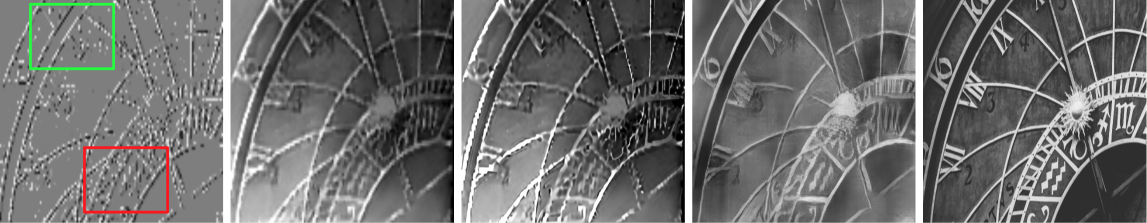}
    \vspace{-3 pt}
    \begin{tabularx}{\linewidth}{
    >{\centering}X
    >{\centering}X
    >{\centering}X
    >{\centering\arraybackslash}X
    }
    \scriptsize EV &  \scriptsize EV+SR $2\scriptstyle\times$  &  \scriptsize Ours $2\scriptstyle\times$ & \scriptsize APS\\
    \end{tabularx}
    \includegraphics[width=1\linewidth]{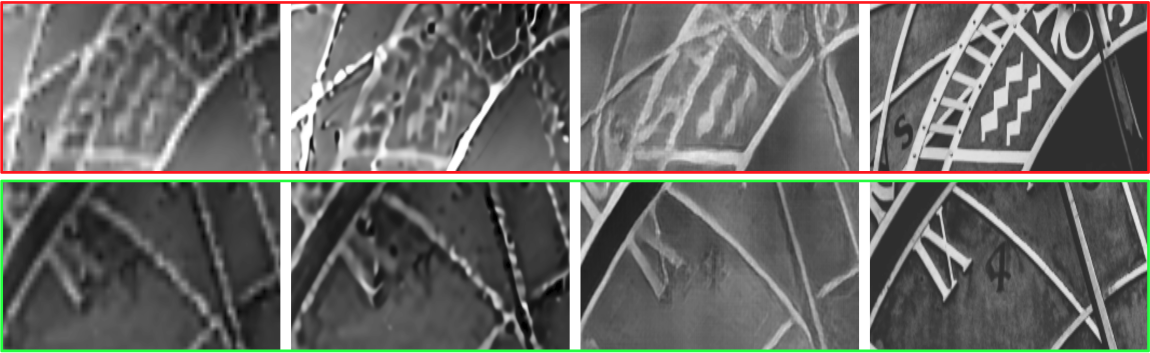}
\caption{Additional comparison between direct event to SR intensity (ours) and event to image to SR intensity in a hierarchical manner (EV+MISR) on simulated sequences. (In addition to the Fig. 5 of main manuscript)}
\label{fig:clock}
    \begin{tabularx}{\linewidth}{
    >{\centering}X
    >{\centering}X
    >{\centering}X
    >{\centering\arraybackslash}X
    }
    \scriptsize EV &  \scriptsize EV $2\scriptstyle\times$  &  \scriptsize Ours $2\scriptstyle\times$ & \scriptsize APS\\
    \end{tabularx}
\includegraphics[width=1\linewidth]{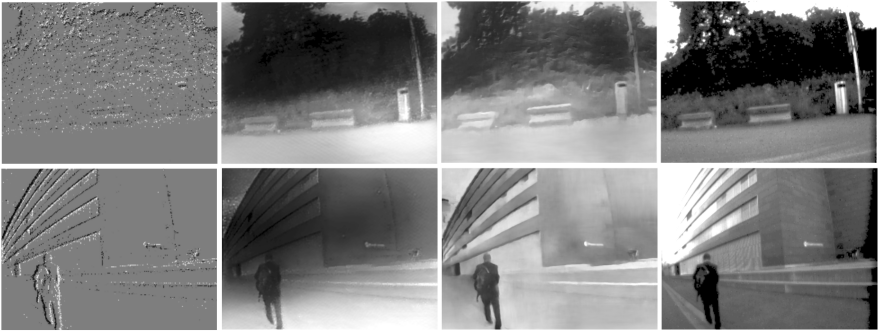}
\caption{Intensity reconstruction in the presence of background noise from far away objects. (In addition to the Fig. 7 of main manuscript)}
\label{fig:outdoor}
\end{figure*}
\begin{figure*}[t!]
    \begin{tabularx}{\linewidth}{
    >{\centering}X
    >{\centering}X
    >{\centering}X
    >{\centering\arraybackslash}X
    }
    \scriptsize Events \hspace{20 pt} & \scriptsize EV \hspace{20 pt} &  \scriptsize Ours $2\scriptstyle\times$ \hspace{20 pt} &  \scriptsize APS \\
    \end{tabularx}
        \includegraphics[width=1\linewidth]{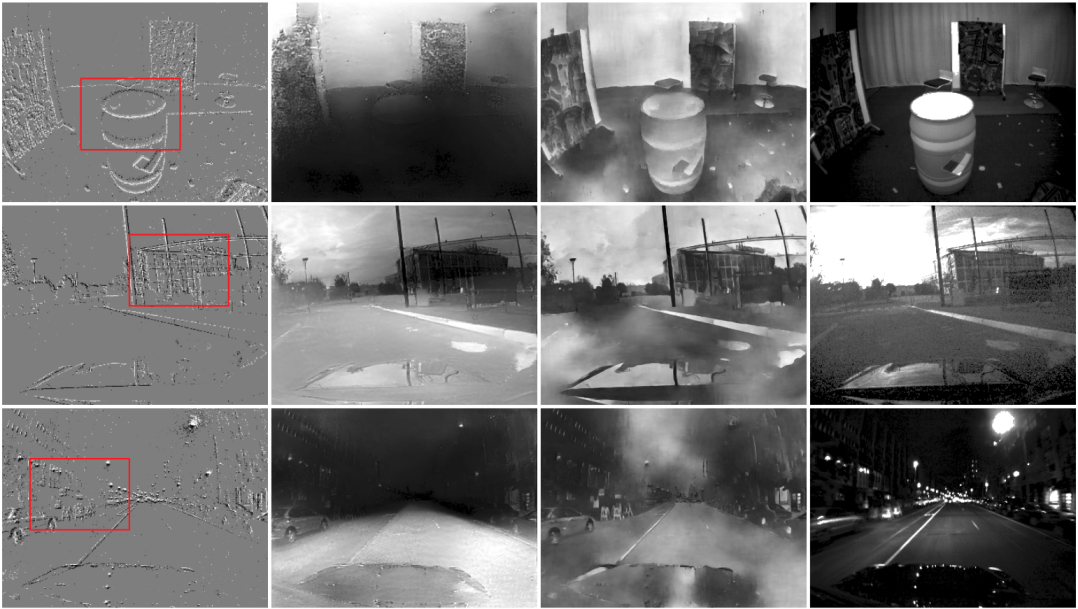}
        
\caption{Expressing the robustness of our intensity image reconstruction in challenging scenes.  
When testing on diverse indoor and outdoor driving scenes with different lighting conditions \cite{zhu2018multivehicle}, our method synthesizes more details while producing less artifacts in comparison to EV and the APS. Please zoom in and compare the suggested regions.}
\label{fig:challenging}
\centering
    \begin{tabularx}{\linewidth}{
    >{\centering}X
    >{\centering}X
    >{\centering}X
    >{\centering\arraybackslash}X
    }
    \scriptsize $t_0$  & \scriptsize $t_1$  &  \scriptsize $t_2$  &  \scriptsize $t_3$\\
    \end{tabularx}
        \includegraphics[width=1\linewidth]{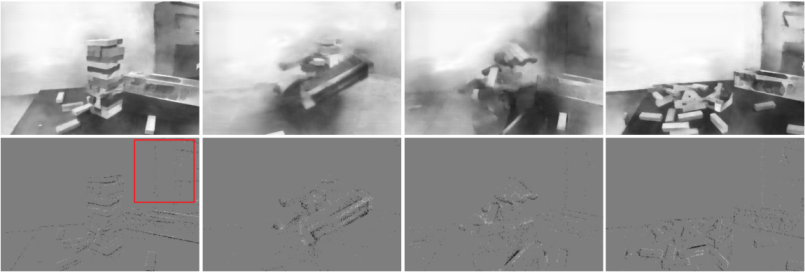}
\caption{Forgetting background details with SBN in rapid object movement.}
\label{fig:limitation}
\end{figure*}

\begin{figure*}[t!]
\centering
    \begin{tabularx}{\linewidth}{
    >{\centering}X
    >{\centering}X
    >{\centering}X
    >{\centering}X
    >{\centering}X
    >{\centering}X
    >{\centering\arraybackslash}X
    }
    \scriptsize APS  & \scriptsize Ours 1K  &  \scriptsize Ours 5K  &  \scriptsize Ours 10K & \scriptsize Events 1K  & \scriptsize Events 5K  &  \scriptsize Events 10K\\
    \end{tabularx}
        \includegraphics[width=1\linewidth]{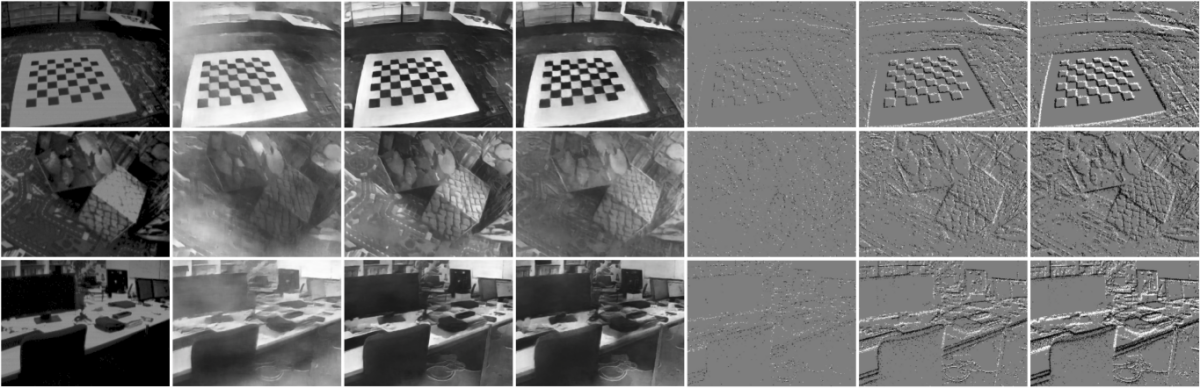}
        \newline
        \vspace{-13 pt}
        \includegraphics[width=1\linewidth]{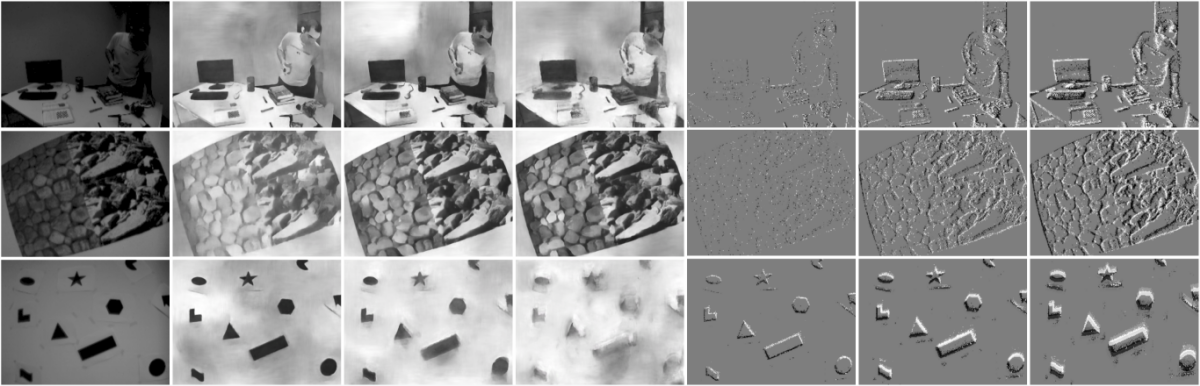}
\caption{Effect of the number of events on reconstruction quality. APS frame shown is for reference.}
\label{fig:1_5_10}
\end{figure*}

\begin{figure*}[t!]
    \begin{tabularx}{\linewidth}{
    >{\centering}X
    >{\centering}X
    >{\centering}X
    >{\centering\arraybackslash}X
    }
    \scriptsize $t_0$  & \scriptsize $t_1$  &  \scriptsize $t_2$  &  \scriptsize $t_3$\\
    \end{tabularx}
\includegraphics[width=1\linewidth]{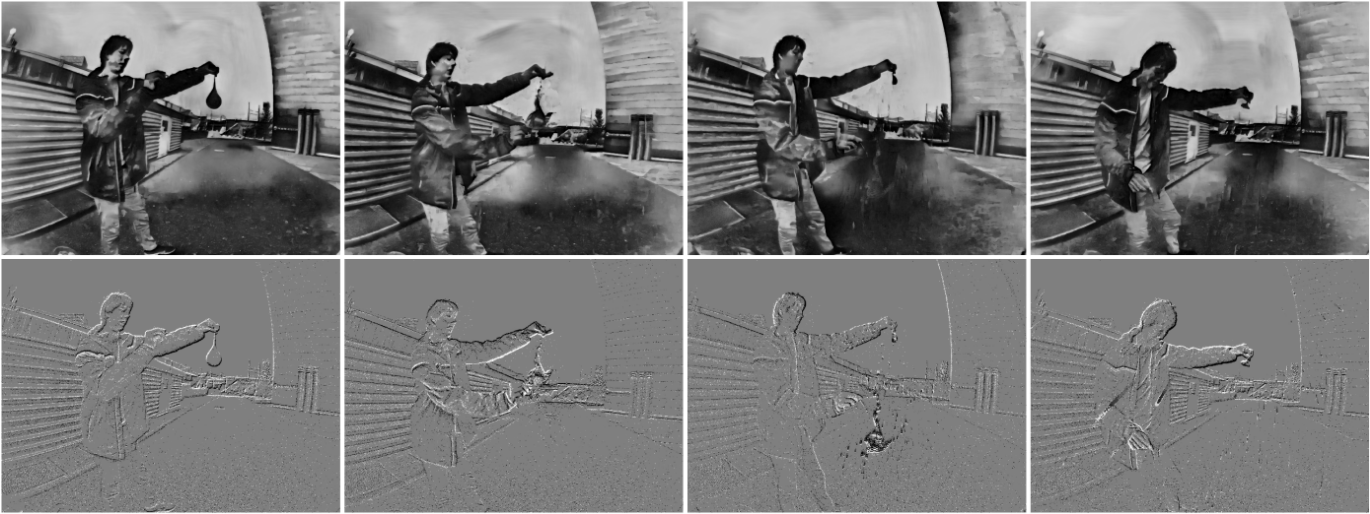}
\caption{A challenging high-speed scenario of popping a water balloon over time ($t_0$ to $t_3$). The intensity details are available in SR dimensions. The background is well reconstructed and the fast moving foreground has been also reconstructed.}
\label{fig:popping_water_balloon}
\end{figure*}

\end{document}